\pdfoutput=1
\def\llm{95}
\def\name{COLE}

\documentclass[11pt]{article}
\usepackage[table,dvipsnames]{xcolor}

\usepackage{acl}

\usepackage{times}
\usepackage{latexsym}

\usepackage[T1]{fontenc}
\usepackage[utf8]{inputenc}

\usepackage{placeins}

\usepackage{microtype}
\usepackage{hyperref}
\usepackage{booktabs}
\usepackage{graphicx}
\usepackage{multirow}
\usepackage{float}
\usepackage{adjustbox}
\usepackage{amsmath}
\usepackage{amssymb}
\usepackage{amsthm}
\usepackage{bm} % Grec en grad avec \bm
\usepackage{mathtools, nccmath}
\usepackage{enumitem}

\usepackage{subcaption}

\usepackage[english]{babel}
\addto\extrasenglish{%

}

\usepackage{csquotes}

\usepackage{tikz}
\usetikzlibrary{shapes.misc, positioning, decorations.pathreplacing, calc}
\usepackage{tcolorbox}
\definecolor{azure}{rgb}{0.0, 0.5, 1.0}
\definecolor{tgb1}{rgb}{0.2666, 0.466, 0.666}
\definecolor{tgb2}{rgb}{0.4, 0.8, 0.9333}
\definecolor{tgb3}{rgb}{0.1333, 0.5333, 0.2}
\definecolor{tgb4}{rgb}{0.8, 0.733, 0.2666}
\definecolor{tgb5}{rgb}{0.9333, 0.4, 0.466}
\definecolor{tgb6}{rgb}{0.666, 0.2, 0.466}
\definecolor{Green}{RGB}{76, 119, 59}

\usepackage{soul}
\DeclareRobustCommand{\hlcbluefonce}[1]{{\sethlcolor{tgb1}\hl{#1}}}
\DeclareRobustCommand{\hlcyellowfonce}[1]{{\sethlcolor{tgb4}\hl{#1}}}

\newcommand{\amine}[1]{\textcolor{red}{(MA)}}

\usepackage{pgf}

\usepackage{collcell}
\usepackage{times}

\usepackage{expex}
\usepackage[linguistics]{forest}
\usepackage{extdash} % needed for suppressing linebreaks at hyphens in a table

\newcommand*{\MinNumber}{0.0}%
\newcommand*{\MidNumber}{60.0} %
\newcommand*{\MaxNumber}{100.0}%

\definecolor{Gray}{gray}{0.9}
\definecolor{cb-blue-green} {RGB}{ 0,  073,  073}
\definecolor{cb-green-sea}  {RGB}{ 0, 146, 146}
\definecolor{cb-rose}       {RGB}{255, 109, 182}
\definecolor{cb-salmon-pink}{RGB}{255, 182, 119}
\definecolor{cb-purple}     {RGB}{ 73,   0, 146}
\definecolor{cb-blue}       {RGB}{ 0, 109, 219}
\definecolor{cb-lilac}      {RGB}{182, 109, 255}
\definecolor{cb-blue-sky}   {RGB}{109, 182, 255}
\definecolor{cb-blue-light} {RGB}{182, 219, 255}
\definecolor{cb-burgundy}   {RGB}{146,   0,   0}
\definecolor{cb-brown}      {RGB}{146,  73,   0}
\definecolor{cb-clay}       {RGB}{219, 209,   0}
\definecolor{cb-green-lime} {RGB}{ 36, 255,  36}
\definecolor{cb-yellow}     {RGB}{255, 255, 109}
\definecolor{cb-grey}       {RGB}{233, 233, 233}

\newcommand{\ApplyGradient}[1]{%
        \ifdim #1 pt > \MidNumber pt
            \pgfmathsetmacro{\PercentColor}{max(min(100.0*(#1 - \MidNumber)/(\MaxNumber-\MidNumber),100.0),0.00)} %
            \hspace{-0.33em}\colorbox{SeaGreen!\PercentColor!Goldenrod!50}{#1}
        \else
            \pgfmathsetmacro{\PercentColor}{max(min(100.0*(\MidNumber - #1)/(\MidNumber-\MinNumber),100.0),0.00)} %
            \hspace{-0.33em}\colorbox{Red!\PercentColor!Goldenrod!50}{#1}
        \fi
}

\newcommand*{\MinNumberTwo}{-70}%
\newcommand*{\MidNumberTwo}{-30} %
\newcommand*{\MaxNumberTwo}{20}%

\newcommand{\ApplyGradienttwo}[1]{%
        \ifdim #1 pt > 0 pt
            \pgfmathsetmacro{\PercentColor}{max(min(100.0*(#1 - \MidNumberTwo)/(\MaxNumberTwo-\MidNumberTwo),100.0),0.00)} %
            \hspace{-0.33em}\colorbox{SeaGreen!\PercentColor!Goldenrod!50}{#1}
        \else
            \pgfmathsetmacro{\PercentColor}{max(min(100.0*(\MidNumberTwo - #1)/(\MidNumberTwo-\MinNumberTwo),100.0),0.00)} %
            \hspace{-0.33em}\colorbox{Red!\PercentColor!Goldenrod!50}{#1}
        \fi
}

\newcolumntype{R}{>{\collectcell\ApplyGradient}c<{\endcollectcell}}
\newcolumntype{D}{>{\collectcell\ApplyGradienttwo}c<{\endcollectcell}}

\title{\name{}: a Comprehensive Benchmark for French Language Understanding Evaluation}

\author{David Beauchemin$^\dagger$, Yan Tremblay$^\dagger$, Mohamed Amine Youssef$^\dagger$ \and Richard Khoury\\
Group for Research in Artificial Intelligence of Laval University (GRAIL)\\
Université Laval, Québec, Canada \\
\texttt{david.beauchemin@ift.ulaval.ca},
\texttt{yan.tremblay.6@ulaval.ca},\\
\texttt{mohamed-amine.youssef.1@ulaval.ca}, 
\texttt{richard.khoury@ift.ulaval.ca}
}

\begin{document}
\maketitle
\begin{abstract}
To address the need for a more comprehensive evaluation of French Natural Language Understanding (NLU), we introduce COLE, a new benchmark composed of 23 diverse task covering a broad range of NLU capabilities, including sentiment analysis, paraphrase detection, grammatical judgment, and reasoning, with a particular focus on linguistic phenomena relevant to the French language. We benchmark \llm{} large language models (LLM), providing an extensive analysis of the current state of French NLU. Our results highlight a significant performance gap between closed- and open-weights models and identify key challenging frontiers for current LLMs, such as zero-shot extractive question-answering (QA), fine-grained word sense disambiguation, and understanding of regional language variations. We release COLE as a public resource to foster further progress in French language modelling.
\end{abstract}

\section{Introduction}
\label{sec:intro}
% \blfootnote{Authors contributed equally to this work.}

The field of Natural Language Processing (NLP) has seen significant progress in recent years, primarily driven by the development of LLMs. 
These models, trained on vast amounts of text data, have achieved state-of-the-art results on a wide range of tasks. Examples can be found in healthcare, where they assist in clinical diagnosis \cite{thirunavukarasu2023large}, and in insurance, where they explain insurance contracts \cite{beauchemin2024quebec}.
To accurately gauge their capabilities and limitations, the community relies on comprehensive benchmarks. 
In the English-speaking world, corpora of tasks like the General Language Understanding Evaluation benchmark (GLUE) \citep{wang-etal-2018-glue}, a suite of nine tasks designed to assess a model's ability to understand natural language (NL), have become the standard for evaluating model competency across a diverse set of NLU tasks.
Following the success of GLUE, similar initiatives have emerged for other languages, such as the benchmark FLUE in French \citep{le2020flaubert} or CLUE in Chinese \citep{xu-etal-2020-clue}.
% The first proposal presents six tasks, ranging from word disambiguation to sentiment classification, whereas the latter proposes ten diverse tasks, including idiom comprehension and question-answering (QA).
These benchmarks have been instrumental in advancing NLP research in various languages by providing language-specific evaluation sets, which make it possible to quantify the development of language-specific resources, like the French LM FlauBERT \citep{le2020flaubert}.
%, and Chinese with  ERNIE 3.0 \citep{sun2021ernie}, by providing standardized evaluation sets

Despite the significant contributions of FLUE, it caused a gap in the evaluation of NLU in French.
Indeed, the benchmark lacks task diversity.
It does not include typical NLU tasks, such as textual entailment (TE), idiom comprehension, QA, linguistic phenomena, paraphrase detection and sentiment analysis.
Thus, there remains a need for a broader set of NLU tasks to evaluate the competency of LMs. 
% Moreover, the FLUE corpus size per task is insufficient in the era of LLM. 
% Thus, a benchmark with a broader set of tasks and a larger corpus size would enable a more granular and robust assessment of an LLM's NLU competency in French. 
Moreover, it should address the unique linguistic challenges endemic to the French language, such as its rich morphology \cite{gross1984lexicon}, grammatical gender \cite{rowlett2007syntax}, and complex syntactic structures \cite{abeille2002syntactic}, thereby providing a more rigorous and nuanced measure of a model's competency.

To address this gap, we introduce the \textbf{CO}rpus for \textbf{L}angue understanding \textbf{E}valuation (\name{}), a French NLU benchmark of 23 tasks. %\footnote{\textit{Colle} is the French translation of \enquote{glue}.}
\name{} is designed to provide a comprehensive and challenging evaluation suite for LLMs, with a focus on a wide variety of tasks, including linguistic phenomena and reasoning types that are particularly relevant to the French language. 
%% Review version
%Our main contributions are as follows:
%\begin{enumerate}[leftmargin=*, noitemsep, topsep=0ex]
%	\item We propose a comprehensive evaluation suite for French NLU, carefully curated to cover a wide range of tasks and corpus size\footnote{Link removed for double-anonymized anonymity.}.
%	\item We benchmark and analyze \llm{} LLMs on \name{}.
%\end{enumerate}

release version
 Our main contributions are as follows:
 \begin{enumerate}[leftmargin=*, noitemsep, topsep=0ex]
	     \item We propose a \href{https://huggingface.co/datasets/graalul/COLE-public}{comprehensive evaluation suite} for French NLU, carefully curated to cover a wide range of tasks and corpus size\footnote{\href{https://huggingface.co/datasets/graalul/COLE-public}{https://huggingface.co/datasets/graalul/COLE-public}}.
	     \item We benchmark and analyze \llm{} LLMs on \name{}.
	 \end{enumerate}

This paper is organized as follows: \autoref{sec:rel_work} presents the related work, then \autoref{sec:cole} describes our tasks. 
\autoref{sec:experiment} details the experimentation setup and \autoref{sec:results} discusses the results. Finally, \autoref{sec:conclusion} concludes the paper and our future work.

\section{Related Work}
\label{sec:rel_work} 
Historically, evaluation of LMs has been conducted using metrics or benchmark corpora \cite{chang2023survey}.
The first approach relies either on task-agnostic metrics, such as perplexity \cite{jelinek1977perplexity}, which measures the quality of the probability distribution of words generated by a model, or on task-specific metrics, like the BLEU score that evaluates a model’s performance for machine translation \cite{Papineni02bleu:a} or MeaningBERT that evaluates meaning preservation in text simplification \cite{beauchemin2023meaningbert}.
However, these metrics do not provide a comprehensive assessment of a model's general NLU competencies.
Task-specific metrics are narrowly-focused by definition, and task-agnostic metrics often fail to capture crucial aspects of language \citep{bowman-dahl-2021-will}.

The second approach relies on the use of large benchmark corpora designed for NLU or other downstream tasks. 
For example, the GLUE benchmark is designed to assess a model's NLU performance on nine distinct tasks. %tasks such as semantic similarity and linguistic acceptability judgment.
%It is composed of nine distinct tasks designed to assess a model's NLU competency. 
These tasks can be grouped into three main categories. 
The first involves single-sentence understanding, and includes a sentences' grammatical acceptability (CoLA) task \citep{warstadt-etal-2019-neural} and a binary sentiment classification task (SST) \citep{socher-etal-2013-recursive}. 
The second category focuses on similarity and paraphrase detection, It features the Microsoft Research Paraphrase Corpus \citep{dolan-brockett-2005-automatically} to identify if two sentences are semantically equivalent, the Quora Question Pairs \citep{wang-etal-2018-glue} to detect duplicate questions, and the Semantic Textual Similarity (STS) task \citep{cer-etal-2017-semeval}, which involves predicting a similarity score between 1 and 5. 
The final category, Natural Language Inference (NLI), which is a cornerstone of NLP evaluation efforts and a fundamental task for NLU. 
NLI, also known as recognizing TE, is the task of determining whether a \enquote{hypothesis} is true (entailment), undetermined (neutral), or false (contradiction) given a \enquote{premise}. 
This task is particularly challenging as it requires a deep understanding of semantic relationships, contextual nuances, and reasoning. 
A model that performs well on NLI typically demonstrates a sophisticated grasp of language that goes beyond surface-level pattern matching, making it a crucial component for many downstream applications such as summarization \citep{bowman-etal-2015-large}.
In GLUE, this category of tasks comprises four distinct challenges: 
the Multi-Genre NLI (MNLI) task \citep{williams-etal-2018-broad}, a large-scale task with both in-domain and cross-domain test sets; the Question NLI task \citep{rajpurkar-etal-2016-squad}, as a binary classification task;
the Recognizing TE (RTE) task \citep{dagan2005pascal}, a binary NLI task; and the Winograd NLI task \citep{levesque2012winograd}, a coreference resolution NLI task.

Following this paradigm, the FLUE benchmark \citep{le2020flaubert} aggregates six tasks to assess LM competency on French, which can also be grouped into three categories. 
The first is text classification, represented by the Cross-Lingual Sentiment (CLS) task, which evaluates sentiment classification on product reviews \citep{prettenhofer2010cross}. 
The second category focuses on sentence-pair understanding, and features a paraphrasing task using the PAWS-X corpus \citep{pawsx2019emnlp} to identify semantic equivalence, and a cross-lingual NLI (XNLI) task to determine the logical relationship between a premise and a hypothesis \citep{conneau-etal-2018-xnli}. 
Finally, the benchmark has a category to probe linguistic and semantic knowledge with a dependency parsing task, designed to analyze grammatical structure \citep{abeille-etal-2003-french, seddah2013overview}, and two Word Sense Disambiguation (WSD) tasks that require identifying the correct meaning of an ambiguous noun or verb in context \citep{segonne2019using}.
%However, the corpus lacks task diversity and is relatively small.

Similarly, the CLUE benchmark \citep{xu-etal-2020-clue} provides a suite of ten tasks designed for Mandarin Chinese. 
It includes two text classification tasks: TNEWS for short news articles and IFLYTEK for longer app descriptions \citep{xu-etal-2020-clue}. 
For sentence-pair understanding, it features the Ant Financial Question Matching Corpus \citep{xu-etal-2020-clue}, which requires identifying semantically equivalent questions, and the Chinese Multi-Genre NLI Task \citep{xu-etal-2020-clue}. 
It also emphasizes reading comprehension, with two distinct extractive QA datasets: CMRC2018 \citep{cui-etal-2019-span} and DRCD \citep{shao-etal-2018-drcd}. Finally, CLUE probes more complex and specialized reasoning through unique tasks: the Winograd Schema Challenge (WSC) for commonsense pronoun resolution \citep{xu-etal-2020-clue}, the Chinese idiom dataset for cloze test idiom understanding \citep{zheng-etal-2019-chid}, the C3 dataset for multi-choice cloze tests requiring causal reasoning \citep{sun-etal-2020-c3}, and the Chinese scientific literature task for abstract-keyword relevance.

% More recently, efforts have been made to create benchmarks for low-resource languages, enabling better evaluation of multilingual models. For instance, BelarusianGLUE \citep{aparovich2025belarusianglue} was introduced as the first NLU benchmark for Belarusian, an East Slavic language classified as vulnerable. 
% It includes five expert-crafted datasets with tasks formulated as binary classification: sentiment analysis, linguistic acceptability, word in context, WSC, and TE.

\section{\name{} Benchmark}
\label{sec:cole}
In this section, we present the 23 tasks that compose \name{} in \autoref{sec:tasks}, divided into three categories: single-sentence, similarity and paraphrasing, and inference. \autoref{table:cole_summary} presents a summary of all our tasks, including instance type, metric, and corpus size. We will then present the evaluation metric and our composite score in \autoref{sec:metric}.

\subsection{Tasks}
\label{sec:tasks}
%We present a summary table of all our tasks, instance type, metric, and corpus size in \autoref{table:cole_summary}, and we describe each task in the following subsections.
\begin{table*}
    \centering
    \resizebox{\textwidth}{!}{%
        \begin{tabular}{llllrrr}
        \toprule
        \textbf{Task Name} & \textbf{Task Type} & \textbf{Instance Type} & \textbf{Metric} & \textbf{Train} & \textbf{Dev} & \textbf{Test} \\
        \midrule
        Allociné & Sentiment Analysis & Sentence & Accuracy & 160,000 & 20,000 & 20,000 \\
        DACCORD & Paraphrase Detection & Sentence Pair & Accuracy & -- & -- & 1,034 \\
        FQuAD & Extractive QA & Context and Question & EM/F1 & -- & 100 & 400 \\
        Fr-BoolQ & Boolean QA & Context and Question & Accuracy & -- & -- & 178 \\
        FraCaS & NLI & Sentence Pair & Accuracy & -- & -- & 346\\
        GQNLI-Fr & NLI & Sentence Pair & Accuracy & 243 & 27 & 30 \\
        LingNLI-Fr & NLI & Sentence Pair & Accuracy & 29,985 & -- & 4,893\\
        MMS & Sentiment Analysis & Sentence & Accuracy & 132,696 & 14,745 & 63,190\\
        MNLI-9/11-Fr & NLI & Sentence Pair & Accuracy & -- & -- & 2,000 \\
        MultiBLiMP-Fr & Grammatical Acceptability & Sentence Pair & Accuracy & 160 & 18 & 77\\
        PAWS-X & Paraphrase Detection & Sentence Pair & Accuracy & 49,401 & 2,000 & 2,000 \\
        PIAF & Extractive QA & Context and Question & EM/F1 & 3,105 & 346 & 384 \\
        QFrBLiMP & Grammatical Acceptability & Sentence Pair & Accuracy & 1,108 & 124 & 529 \\
        QFrCoLA & Grammatical Acceptability & Sentence & Accuracy & 15,846 & 1,761 & 7,546 \\
        QFrCoRE & Definition Matching & List of Sentences & Accuracy & -- & -- & 4,633 \\
        QFrCoRT & Definition Matching & List of Sentences & Accuracy & -- & -- &  201 \\
        RTE3-Fr  & NLI & Sentence Pair & Accuracy & -- & 800 & 800 \\
        SICK-Fr & NLI & Sentence Pair & Accuracy & 4,439 & 495 & 4,906 \\
        STS22 & STS & Sentence Pair & Accuracy & 101 & -- & 72 \\
        Wino-X-LM & Pronoun Resolution & Sentence & Accuracy & -- & -- & 2,793 \\
        Wino-X-MT & Pronoun Resolution  & Sentence and Translation & Accuracy & -- & -- & 2,988 \\
        WSD-Fr & WSD & Sentence & EM & 269,821 & -- & 3,121\\
        XNLI-Fr & NLI & Sentence Pair & Accuracy & 393,000 & 2,490 & 5,010 \\
        \bottomrule
        \end{tabular}%
    }
    \caption{\name{}'s 23 tasks summary with instance types, evaluation metrics, and sizes per split.}
    \label{table:cole_summary}
    \vspace{-1.5em}
\end{table*}

\subsubsection{Single-Sentence Tasks}
\begin{enumerate}[leftmargin=*, noitemsep, topsep=0ex]
    \item \textbf{Allociné} \cite{blard2019allocine} is a task that focuses on sentiment analysis using movie reviews scraped from the Allociné website. 
    Each instance is a single sentence expressing a user-generated opinion. 
    The model must classify the sentiment as negative (0) or positive (1). This task evaluates the ability of LLMs to understand sentiment in informal language.
    \item \textbf{MMS-Fr} \cite{augustyniak2023massively} is a sentiment analysis task using the French subset of the Massive Multilingual Sentiment (MMS) corpus. 
    Each instance is a single text entry, and the model must classify its sentiment into one of three categories: negative (0), neutral (1), or positive (2). 
    This task evaluates an LLM's ability to discern sentiment across the various domains and text sources included in the original collection. 
    \item \textbf{QFrCoLA} \cite{beauchemin2025qfrcolaquebecfrenchcorpuslinguistic} is a grammatical acceptability classification task for Quebec-French. Each instance is labelled as unacceptable (0) or acceptable (1). The model must classify whether the sentence is grammatically acceptable or not. 
    This task assesses the grammar competency of an LLM.
\end{enumerate}

\subsubsection{Similarity and Paraphrase Tasks}
\begin{enumerate}[leftmargin=*, noitemsep, topsep=0ex]
    \item \textbf{DACCORD} \cite{skandalis-etal-2024-new-datasets} is a paraphrase detection task between sentence pairs. Each instance consists of two sentences, and the model must determine whether they convey the same meaning (0) or contradict each other (1). 
    The dataset contains sentence pairs manually curated to reflect NL use and covers a variety of topics, including political discourse. 
    % This makes it more challenging and domain-relevant than generic paraphrase datasets. 
    This task assesses an LLM's ability to comprehend paraphrasing across a wide range of topics.
    \item \textbf{FQuAD} \cite{dhoffschmidt-etal-2020-fquad} is a French extractive QA task built from Wikipedia articles. 
    Given a context paragraph and a question in French, the model must extract a contiguous span of text from the paragraph that answers the question. 
    The dataset is designed to evaluate the ability of models to understand and retrieve factual information in French. 
    \item \textbf{Fr-BoolQ} \cite{french_boolq} is a binary QA task translated into French from the original BoolQ dataset \cite{clark2019boolq}.
    Each instance consists of a short context and a yes/no question. 
    The model must determine whether the context supports the answer to the question. 
    The questions are naturally occurring and not guaranteed to be answerable, making the task challenging. 
    The goal is to predict \enquote{yes} (1) if the context entails the answer, and \enquote{no} (0) otherwise. 
    \item \textbf{PAWS-X} \cite{pawsx2019emnlp} is a paraphrase detection task designed to evaluate models' ability to detect whether two sentences in French convey the same meaning despite having different surface forms. 
    Each instance presents a pair of sentences that are often lexically similar but semantically distinct. 
    This task assesses an LLM's competency in understanding semantics.
    \item \textbf{PIAF} \cite{keraron-EtAl:2020:LREC} is a French extractive QA task developed from government and public-domain documents. 
    Given a context passage and a question in French, the model must extract the precise span of text that answers the question. 
    PIAF is designed to evaluate models in realistic information access settings, with questions sourced from real user needs and verified by human annotators. 
    It complements FQuAD by covering a broader range of topics relevant to public services. 
    \item \textbf{QFrCoRE} and \textbf{QFrCoRT} \cite{qfrcore} are, respectively, an expression and a regional terms definition matching task.
    Each example consists of a local Quebec idiom or word and a list of ten potential definitions for the instance.
    The model must determine the appropriate definition from the candidate list.
    This task assesses an LLM's capacity to comprehend local expressions or regional terms.
    \item \textbf{STS22} \cite{10.5555/2387636.2387697} is an STS task that evaluates the degree of semantic equivalence between pairs of French sentences. 
    Each input consists of two sentences, and the model must assign an integer-valued similarity score ranging from 1 (completely unrelated) to 4 (identical meaning). 
    This task evaluates an LLM's ability to understand semantic equivalences.
\end{enumerate}

\subsubsection{Inference Tasks}
\begin{enumerate}[leftmargin=*, noitemsep, topsep=0ex]
    \item \textbf{FraCaS} \cite{richard2024FraCaS} is an NLI task. 
    The dataset is designed to evaluate a model's semantic reasoning capabilities across a broad and structured range of linguistic phenomena, such as quantifiers, plurality, anaphora, and ellipsis. 
    \item \textbf{GQNLI-Fr} \cite{skandalis-etal-2024-new-datasets} is an NLI task that uses an automatically-generated French translation of the English GQNLI dataset \cite{cui-etal-2022-generalized}. 
    It focuses specifically on generalized quantifiers (e.g. some, all, none, few) and tests LLMs' ability to reason over these constructs.  
    \item \textbf{LingNLI-Fr} \cite{skandalis-etal-2024-new-datasets} is an NLI task that uses an automatically-generated French translation of the English LingNLI dataset \cite{parrish-etal-2021-putting-linguist}. 
    \item \textbf{MNLI-9/11-Fr} \cite{N18-1101} is an NLI task based on a French-translated subset of the MultiNLI dataset, focusing on sentence pairs on the topic of 9/11. 
    This task assesses an LLM's ability to reason about hypothesis understanding in NL.  
    \item \textbf{MultiBLiMP-Fr} \cite{jumelet2025multiblimp} is a grammatical judgment task utilizing the French subset of the Multilingual Benchmark of Linguistic Minimal Pairs (MultiBLiMP). 
    Each instance consists of a minimal pair of sentences: one grammatically correct and one incorrect, differing by a single, targeted linguistic feature. 
    The model is required to identify the grammatically-acceptable sentence from the pair. 
    This task evaluates the model's knowledge of six linguistic phenomena, including syntax, morphology, and agreement.
    \item \textbf{QFrBLiMP} \cite{beauchemin2025qfrblimpquebecfrenchbenchmarklinguistic} is a grammatical judgment task for Quebec French.
    The model chooses which of two sentences is grammatically correct. 
    This task assesses the LLM's grammatical competency across 20 linguistic phenomena.
    \item \textbf{RTE3-Fr} \cite{skandalis-etal-2024-new-datasets} is an NLI task based on a French translation of the RTE3 dataset \cite{giampiccolo2007third}. 
    It is designed to test fine-grained reasoning over short texts. 
    \item \textbf{SICK-Fr} \cite{sickfr} is an NLI task derived from the French version of SICK dataset \cite{bentivogli2016sick}. 
    This dataset tests the LLM over a broad range of linguistic phenomena, including negation, and paraphrasing. %compositional semantics,
    \item \textbf{Wino-X-LM} \cite{emelin-sennrich-2021-wino} is a pronoun resolution task in the form of coreference resolution. 
    Each example presents a French sentence containing an ambiguous pronoun and two possible antecedents. 
    The model must select the correct referent based on context, thereby evaluating its ability to resolve gender and number agreement in pronominal references. 
    %The task is to test the LM's ability to understand ambiguous pronouns and disambiguate them.
    \item \textbf{Wino-X-MT} \cite{emelin-sennrich-2021-wino} is a natural pronoun resolution task.
    Given two French translations of an original English sentence, each differing only in the gender of a pronoun, the model must choose the translation that correctly resolves the referent according to the context. 
    It tests the model’s sensitivity to grammatical and semantic alignment across languages. 
    \item \textbf{WSD-Fr} \cite{le2020flaubert} is a WSD task from the FLUE benchmark that evaluates a model's ability to identify the correct meaning of an ambiguous verb in a given context. 
    Each instance consists of a sentence, and the model must select the verb/noun that is ambiguous.
    \item \textbf{XNLI-Fr} \cite{conneau-etal-2018-xnli} is an NLI task based on the French subset of the XNLI corpus, which extends the MultiNLI dataset to 15 languages \cite{N18-1101}. 
    This task assesses LM's ability to reason about hypothesis understanding in NL.
\end{enumerate}

\subsection{Evaluation Metrics}
\label{sec:metric}

\subsubsection{Task Evaluations}
To evaluate competency, we utilize task-specific evaluation metrics that are aligned with the original protocol. All three metrics are in the $[0, 1]$ range.
%Below, we describe the metrics used across \name{}.

\begin{enumerate}[leftmargin=!, noitemsep, topsep=0ex, labelindent=0pt,  wide=0pt]
    \item[]\textbf{Accuracy} measures the proportion of correct predictions over the total number of instances.
    \item[]\textbf{Exact Match} (EM) \cite{wang-etal-2018-glue} evaluates whether the predicted answer exactly matches the reference answer. It is a strict metric where any semantic mismatch results in a zero score. 
    \item[]\textbf{F1 Score} is a measure that assesses the token-level overlap between predicted and reference spans. In contrast to EM, it allows partial credit for semantic mismatches. 
    A prediction that overlaps significantly with the answer still receives a high score, even if it is not an exact match.
\end{enumerate}

\subsubsection{Composite Score}
\label{sec:compositescore}
To evaluate the overall model performance across the diverse tasks in \name{}, we compute a composite score (CS) following a methodology similar to that of GLUE. 
This score is calculated as the unweighted mean of per-task scores, where each task contributes equally, regardless of its size or type. 
For tasks reporting multiple main metrics (e.g. FQUaD), we first average these metrics to obtain a single task score. 
Finally, we multiply the weighted mean by $100$ to obtain a normalized rating in the range of $[0, 100]$, so the final composite score can be expressed as a percentage. 
This provides a single and easily-interpretable value to compare models’ general NLU capabilities in French.

\section{Experiments}
\label{sec:experiment} 
%In this section, we present our evaluation settings in \autoref{sec:eval} and our \llm{} selected benchmarked models and baselines in \autoref{sec:models}.

\subsection{Evaluation Settings}
\label{sec:eval}
We evaluate a diverse set of French LLMs using a zero-shot evaluation setup. 
Each task is framed as an NL prompt that describes it, and the model produces an appropriate output based solely on its pretrained capabilities. 
All LLMs are evaluated on the same shared test set to ensure fair comparison. 
Depending on the nature of the task, models either select a response from a predefined set of labels or generate an answer. 
Evaluation is conducted automatically using task-specific metrics.

\subsection{Models}
\label{sec:models}
\subsubsection{Baseline Models}
We use a \texttt{Random} selection algorithm as our baseline.
For our classification tasks, it randomly selects one of the potential candidates as the answer.
For example, in a binary classification such as \textbf{Fr-BoolQ}, it would  select either $0$ or $1$, while for NLI tasks, such as FraCaS, it selects either $0$, $1$, or $2$.
For the text extraction tasks, namely, FQuAD, PIAF and WSD-Fr, it randomly selects a word from the whitespace-split sentence. 
For example, given the sentence \enquote{I love apples.}, it would whitespace-split it to $[\text{\enquote{I}}, \text{\enquote{love}}, \text{\enquote{apples}}]$, then select randomly a word from that set as the answer.% the word index 2, means we select \enquote{love}. %\footnote{Using a 1-based index counting.}
We use the seed $42$ to facilitate the reproducibility of our results.

\subsubsection{LLM}
To ensure a thorough and representative analysis of the current LM landscape, we selected \llm{} to cover four aspects of LLM specifications:

\begin{enumerate}[leftmargin=*, noitemsep, topsep=0ex]
    \item \textbf{A Mix of Access Paradigms:} We include both closed- and open-weights LLMs. 
    \item \textbf{Variety in Size:} The selected models span a large range of parameter counts, from smaller models (under 1 billion parameters) to the largest closed-weights LLMs available as of mid-2025 (around 100 billion parameters).
    \item \textbf{Variety in Capability:} We intentionally included models marketed as having advanced \enquote{reasoning} capabilities ($\Gamma$) to assess if this specialization translates to better performance.
    \item \textbf{Model Specialization:} We include models optimized for French ($\Upsilon$), to test whether this leads to better performance.
\end{enumerate}

To select the LMs, we leverage two leaderboards: the \href{https://lmarena.ai/leaderboard/text}{Text Arena} and the \href{https://huggingface.co/spaces/open-llm-leaderboard/open_llm_leaderboard#/}{Open LLM} leaderboards.
We present our selected models in \autoref{an:selectedllmdetails}, and our hardware and budget in \autoref{an:hardware}.

\section{Results and Discussion}
\label{sec:results}

In this section, we present and analyze the performance of the \llm{} benchmarked LLMs on \name{}'s 23 tasks. 
We present our CS results in \autoref{tab:results} and our complete results are detailed in \autoref{an:results_complete}. 
We structure our discussion around overall performance trends, the impact of model characteristics such as size and specialization, and a closer look at performance on specific task categories.

\subsection{Overall Performance Analysis}
Our evaluation reveals a wide performance spectrum, with the CS ranging from 28.38\% (\texttt{SmolLM2-135M}) to 70.12\% (\texttt{GPT-5-mini-2025-08-07}). A clear trend emerges from the leaderboard: closed-weights LLMs dominate the top ranks. 
%This dominance is underscored by the fact that the 20 highest-scoring models are all closed-weights LLMs, which consistently outperform the rest of the field, establishing a significant performance gap between them and the open-weights models.
%Indeed, the best-performing open-weights model, \texttt{Qwen3-14B}, achieves a CS of 45.17\%, placing it considerably behind the top-tier closed-weight LLMs. 
Indeed, the top 23 highest-scoring models are all closed-weights LLMs, while the best-performing open-weight model, \texttt{Qwen3-max}, achieves a CS of 49.14\%, 10\% less than the next-best-performing closed-weights model and 20\% less than the best-performing closed-weights model. 
One might assume that this is due to the closed-weight models being larger than open-weight models, since we were limited in the size of open-weights models we could run on our local hardware (see \autoref{an:hardware}) while the closed-weight models were run through API calls.
However, even the \enquote{mini} versions of leading closed-weights models substantially outperform larger open-weights models. Moreover, there is no substantial performance difference between regular and \enquote{mini} versions of closed-weight models. For example, the performance decline from \enquote{Grok-3} to \enquote{Grok-3-mini} is less than one percent, and the \enquote{mini} version of \enquote{GPT-5} actually outperforms its larger version.
This suggests that the vast and high-quality proprietary training datasets,  data quality and extensive post-training alignment procedures of closed-weights LLMs are the determining factors, rather than raw parameter count.

%This gap highlights the current advantage held by closed models, which is likely attributable to a combination of model size, architectural innovations, vast and proprietary training datasets, and extensive post-training alignment procedures.
% However, this disparity is likely influenced by model scale, as the parameter counts of closed-weights models are generally not disclosed but are presumed to be substantially larger than those of open-weights models included in our experiments.

%Interestingly, even the \enquote{mini} versions of leading closed-weights models substantially outperform larger open-weights models, suggesting that data quality and fine-tuning are at least as critical as raw parameter count.
%For example, the performance difference between \enquote{Grok-3} and \enquote{Grok-3-mini} is less than one percent.
%Even in some cases, like \enquote{GPT-5}, the mini version obtains better results than the larger version.
The good performance of smaller closed-weights LLMs compared to their larger counterparts may also be due to the knowledge distillation process \cite{hinton2014distilling}.
Indeed, the \enquote{mini} or \enquote{flash} version of an LLM is typically a distilled model of a larger LLM \cite{gou2021knowledge}.
The massive \enquote{teacher} model (i.e. GPT-5) is used to train the smaller \enquote{student} model (i.e. GPT-5-mini). 
The student learns to replicate the teacher's correct outputs and reasoning patterns, essentially inheriting the most important knowledge in a much more compact form.
This process hyper-focuses the mini LM on core language understanding and instruction-following skills \cite{sanh2019distilbert, ouyang2022training}. %chung2024scaling
The larger LM, in contrast, retains a wider, more general set of capabilities that may not be relevant to the benchmark tasks \cite{liang2022holistic}. % weiemergent

\subsection{The Impact of Model Specialization}
We analyze two types of specialization: models optimized for French ($\Upsilon$) and for reasoning ($\Gamma$).

French-specialized models show competitive but mixed results. 
For instance, \texttt{Chocolatine-2-14B-it-v2.0.3} stands out as the strongest open-weights models with a CS of 45.05. 
Its strong performance on grammatical judgment tasks, such as \textbf{MultiBLiMP-Fr} (94.81\%), suggests a superior grasp of French linguistic structure. 
However, this does not uniformly translate to semantic and reasoning tasks, where general-purpose LLMs surpass it. 
While language-specific pre-training is effective for capturing syntactic and grammatical nuances, it may not be sufficient to bridge the gap in broader NLU capabilities.

Reasoning models generally perform well on inference-heavy tasks. 
LLMs like \texttt{Claude-opus-4} achieve high scores on NLI tasks such as \textbf{LingNLI-Fr} and \textbf{SICK-Fr}. 
However, this result is not consistent. 
Performance on \textbf{FraCaS}, a task designed to probe deep semantic and logical phenomena, remains challenging even for top-tier reasoning models. In fact, the best result in this task (66.27\%) was achieved by \texttt{GPT-4o-mini-2024-07-18}, a non-reasoning model. % achieving the highest score at 66.27\%. 
This highlights that while current reasoning models are adept at pattern-matching forms of inference, they still struggle with more complex, structured logical reasoning.

\subsection{Performance Across Task Categories}
\subsubsection{Areas of Strength}
\paragraph{Semantic Understanding.}
Models generally excel at tasks requiring coarse-grained semantic understanding. 
For example, on sentiment analysis (\textbf{Allociné}) and paraphrase detection (\textbf{DACCORD}), top models consistently achieve accuracy scores above 95\%. 
Similarly, many NLI tasks, such as \textbf{XNLI-Fr}, are handled effectively by the leading models, which score well above 70\%. 
These results suggest that high-level sentence and sentence-pair understanding is a well-developed capability.

\paragraph{Grammatical Judgment.}
Beyond semantic tasks, the LLMs demonstrate a strong competency in French grammatical structure. 
It is most evident in the \textbf{MultiBLiMP-Fr} task, where many leading models achieve near-perfect results, with several scoring a perfect 100\%. 
This high level of performance indicates that the syntactic rules and formal structures of the French language have been robustly learned.
However, this corpus is composed of online text; thus, performance may better reflect proficiency in recognizing common online linguistic patterns rather than a comprehensive grasp of formal grammatical rules.

\subsubsection{Challenging Frontiers}
\paragraph{Extractive QA.} 
Performance in \textbf{FQuAD} and \textbf{PIAF} is low, especially for the EM metric. 
A large number of LLMs, including top-tier ones like \texttt{Claude-opus-4}, achieve a 0\% EM score. 
While their F1 scores are sometimes higher, the inability to extract verbatim text suggests a strong tendency to rephrase or generate answers rather than strictly follow the extraction instruction. 
It highlights a fundamental challenge in adhering to instruction for zero-shot QA.

\paragraph{Regional LU.} 
The regional tasks \textbf{QFrCoRE} and \textbf{QFrCoRT}, which focused on defining Quebec-French expressions and terms, represent a major blind spot. 
Most open-weights LLMs score near the random baseline of 9.1\%, indicating a lack of regional and cultural linguistic knowledge. 
The performance of closed-weights LLMs, such as \texttt{Claude-opus-4}, which scores an exceptional 93.46\% on QFrCoRE and 97.66\% on QFrCoRT, is a significant outlier and demonstrates that capturing such knowledge is possible, though not yet widespread.
This significant gap between closed- and open-weights LLMs suggests that mastering deep cultural and dialectical nuance is not an emergent property of model architecture alone, but rather a direct consequence of the breadth and diversity of the training data. 
The proprietary corpora used by top commercial labs likely contain a far richer concentration of regional web content, literature, and media from Québec than is available in standard open-source datasets. 
The exceptional performance of these few models, therefore, demonstrates that capturing such specific linguistic knowledge is possible.

\paragraph{Word Sense Disambiguation.} 
This task was also challenging, with the vast majority of models failing to produce correct outputs, resulting in scores near 0\%. 
It suggests that while models have a broad contextual understanding, their ability to perform fine-grained lexical disambiguation in a zero-shot manner is severely limited.

\begin{figure*}
    \centering
    \footnotesize
    \begin{tabular}{lclc}
    \toprule
    LLM & \begin{tabular}[c]{@{}c@{}}CS\\ Acc.\\(\%) ($\blacktriangledown$)\end{tabular} &  LLM & \begin{tabular}[c]{@{}c@{}}CS\\ Acc.\\(\%) ($\blacktriangledown$)\end{tabular}\\
    \midrule
    \texttt{GPT-5-mini-2025-08-07} ($\Gamma$) & 70.12 & \texttt{DeepSeek-R1-Distill-Qwen-14B} ($\Gamma$) & 37.77 \\
    \texttt{o3-mini-2025-01-31} ($\Gamma$) & 68.98 & \texttt{Llama-$3.2$-1B} ($\Gamma$) & 37.62 \\
    \texttt{GPT-$4.1$-mini-2025-04-14} & 67.68 & \texttt{Mixtral-8x7B-v$0.1$} & 37.04 \\
    \texttt{Claude-opus-4-20250514} ($\Gamma$) & 67.13 & \texttt{Lucie-7B-it-human-data} ($\Upsilon$) & 35.41 \\
    \texttt{Claude-sonnet-4-20250514} ($\Gamma$) & 65.76 & \texttt{OLMo-2-1124-7B-it} & 35.20 \\
    \texttt{GPT-4o-mini-2024-07-18} & 65.72 & \texttt{Qwen$2.5$-$1.5$B} & 34.88 \\
    \texttt{Gemini-$2.5$-pro} ($\Gamma$) & 65.43 & \texttt{Granite-$3.3$-8b-it} & 34.80 \\
    \texttt{DeepSeek-chat} & 65.20 & \texttt{Qwen$2.5$-$0.5$B} & 34.38 \\
    \texttt{o1-mini-2024-09-12} ($\Gamma$) & 64.44 & \texttt{Phi-$3.5$-mini-it} & 34.28 \\
    \texttt{GPT-$4.1$-2025-04-14} & 64.38 & \texttt{Gemma-2-2b} ($\Gamma$) & 34.22 \\
    \texttt{o1-2024-12-17} ($\Gamma$) & 64.36 & \texttt{OLMo-2-0425-1B-it} & 34.19 \\
    \texttt{o3-2025-04-16} ($\Gamma$) & 64.15 & \texttt{Gemma-2-27b} ($\Gamma$) & 34.19 \\
    \texttt{GPT-5-2025-08-07} ($\Gamma$) & 63.61 & \texttt{OLMo-2-1124-13B-it} & 34.12 \\
    \texttt{GPT-4o-2024-08-06} & 63.04 & \texttt{Chocolatine-14B-it-DPO-v$1.3$} ($\Upsilon$) & 33.95 \\
    \texttt{Grok-3-latest} ($\Gamma$) & 62.34 & \texttt{Gemma-2-9b} ($\Gamma$) & 33.86 \\
    \texttt{Grok-4-0709} & 62.31 & \texttt{SmolLM2-$1.7$B-it} & 33.86 \\
    \texttt{Gemini-$2.5$-flash} & 62.21 & \texttt{Gemma-2-2b-it} ($\Gamma$) & 33.85 \\
    \texttt{DeepSeek-reasoner} ($\Gamma$) & 62.20 & \texttt{OLMo-2-1124-7B} & 33.77 \\
    \texttt{Grok-3-mini-latest} ($\Gamma$) & 62.01 & \texttt{SmolLM2-$1.7$B} & 33.77 \\
    \texttt{GPT-oss-120b} ($\Gamma$) & 61.97 & \texttt{Mixtral-8x7B-it-v$0.1$} & 33.77 \\
    \texttt{Grok-3-fast-latest} ($\Gamma$) & 61.95 & \texttt{DeepSeek-R1-Distill-Qwen-7B} ($\Gamma$) & 33.34 \\
    \texttt{Grok-3-mini-fast-latest} ($\Gamma$) & 61.57 & \texttt{Llama-$3.2$-3B} ($\Gamma$) & 33.26 \\
    \texttt{Mistral-large-latest} ($\Gamma$) & 60.70 & \texttt{Apertus-8B-it-2509} & 33.20 \\
    \texttt{Pixtral-large-latest} & 60.48 & \texttt{Meta-Llama-$3.1$-8B} ($\Gamma$) & 33.12 \\
    \texttt{Qwen-max} & 49.14 & \texttt{OLMo-2-0325-32B} & 33.05 \\
    \texttt{Qwen3-14B} & 45.17 & \texttt{OLMo-2-0425-1B} & 32.84 \\
    \texttt{Chocolatine-2-14B-it-v$2.0$.3} ($\Upsilon$) & 45.05 & \texttt{Gemma-2-27b-it} ($\Gamma$) & 32.81 \\
    \texttt{QwQ-32B} ($\Gamma$) & 44.94 & \texttt{Apertus-8B-2509} & 32.48 \\
    \texttt{DeepSeek-R1-Distill-Qwen-32B} ($\Gamma$) & 44.92 & \texttt{Qwen$2.5$-$1.5$B-it} & 32.32 \\
    \texttt{Qwen3-14B-Base} & 44.49 & \texttt{Qwen$2.5$-3B-it} & 32.25 \\
    \texttt{French-Alpaca-Llama3-8B-it-v$1.0$} ($\Gamma$$\Upsilon$) & 44.42 & \texttt{Llama-$3.2$-1B-it} ($\Gamma$) & 32.15 \\
    \texttt{Qwen$2.5$-14B-it} & 44.01 & \texttt{Qwen$2.5$-3B} & 32.08 \\
    \texttt{S$1.1$-32B} ($\Gamma$) & 42.53 & \texttt{OLMo-2-1124-13B} & 31.61 \\
    \texttt{Phi-4} & 42.16 & \texttt{DeepSeek-R1-Distill-Llama-8B} ($\Gamma$) & 31.51 \\
    \texttt{Granite-$3.2$-8b-it} & 41.33 & \texttt{RandomBaselineModel} & 31.22 \\
    \texttt{Qwen$2.5$-32B} & 41.19 & \texttt{Qwen$2.5$-$0.5$B-it} & 30.74 \\
    \texttt{Qwen$2.5$-7B-it} & 40.19 & \texttt{Aya-23-8b} & 30.71 \\
    \texttt{Meta-Llama-$3.1$-8B-it} ($\Gamma$) & 40.13 & \texttt{Gemma-2-9b-it} ($\Gamma$) & 30.34 \\
    \texttt{Deepthink-Reasoning-14B} ($\Gamma$) & 40.07 & \texttt{CroissantLLMBase} ($\Upsilon$) & 30.21 \\
    \texttt{Deepthink-Reasoning-7B} ($\Gamma$) & 39.96 & \texttt{SmolLM2-135M-it} & 29.84 \\
    \texttt{Qwen$2.5$-14B} & 39.79 & \texttt{SmolLM2-360M} & 29.83 \\
    \texttt{Qwen$2.5$-7B} & 39.78 & \texttt{GPT-oss-20b} ($\Gamma$) & 29.72 \\
    \texttt{Reka-flash-3} ($\Gamma$) & 38.77 & \texttt{Lucie-7B-it-v$1.1$} ($\Upsilon$) & 29.18 \\
    \texttt{Qwen$2.5$-32B-it} & 38.55 & \texttt{SmolLM2-360M-it} & 29.10 \\
    \texttt{Granite-$3.3$-8b-base} & 38.42 & \texttt{DeepSeek-R1-0528-Qwen3-8B} & 29.06 \\
    \texttt{Llama-$3.2$-3B-it} ($\Gamma$) & 38.27 & \texttt{OLMo-2-0325-32B-it} & 28.45 \\
    \texttt{Aya-expanse-8b} & 37.87 & \texttt{SmolLM2-135M} & 28.38 \\
    \texttt{Lucie-7B} ($\Upsilon$) & 37.84 &  &  \\
    \bottomrule
    \end{tabular}
    \caption{Composite score performance of all \llm{} LLMs. Scores are reported as percentages (\%) and are ranked in descending order ($\blacktriangledown$).}
    \label{tab:results}
    % \vspace{-1.5em}
\end{figure*}

% \begin{figure}
%     \centering
%     \includegraphics[width=\linewidth]{figs/model_size_vs_score.png}
%     \caption{Enter Caption}
%     \label{fig:placeholder}
% \end{figure}

\section{Conclusion and Future Works}
\label{sec:conclusion}
In this paper, we introduce \name{}, a comprehensive benchmark for French NLU comprising 23 tasks. 
Our goal was to address a critical gap in the French NLP ecosystem by providing a more challenging and wide-ranging benchmark for LLMs than previously available. 
By benchmarking \llm{} LLMs, we assess the current landscape of NLU capabilities in French.
Our findings reveal a performance gap between closed and open-weights models, with the former consistently achieving superior results. 
This advantage is particularly pronounced in tasks requiring deep understanding of French. 
While models demonstrate strong capabilities in core semantic and grammatical tasks, our benchmark successfully highlights several challenging frontiers for future research, including zero-shot extractive QA and understanding regional linguistic variations.

For future work, we plan to expand COLE to include tasks that cover additional NLU phenomena, such as multi-hop reasoning and dialogue understanding. We also aim to develop a non-public test set to provide a more robust measure of generalization, mitigating the potential for data contamination. 
Our hope is that COLE will serve as a valuable resource for the community, driving the development of more capable and nuanced language models for the French language.

\clearpage
\section*{Limitations}
While we believe \name{} is a significant contribution to the evaluation of French NLU, we acknowledge several limitations that offer avenues for future research.

\paragraph{Data Contamination} A primary limitation of our benchmark is the potential for data contamination. 
All datasets included in \name{} are publicly available. 
Given that LLMs are trained on vast web scrapes, it is highly probable that their training data included portions of the training, validation, and even test sets of these datasets. 
Consequently, our zero-shot evaluation may not purely measure a model's generalization capabilities but could also reflect memorization. 
This is a pervasive challenge in modern LLM evaluation, and holistic benchmarks now emphasize the need to actively track and mitigate such contamination \citep{liang2022holistic, jiang2024investigating}.

\paragraph{Mix of French Varieties} \name{} intentionally incorporates datasets from different varieties of French, notably including several from Québec, namely, QFrCoLA, QFrBLiMP, QFrCoRE, QFrCoRT, alongside datasets primarily based on French from France. 
While this diversity reflects the richness of the language, our composite score does not distinguish between them. 
It is well-documented that NLP models can exhibit performance disparities across dialects, and an aggregated score can obscure these nuances \citep{blodgett-etal-2016-demographic}. 
Future work could involve creating dialect-specific sub-scores to provide a more granular analysis.

\paragraph{Reliance on Machine-Translated Datasets} Several of our inference tasks, such as GQNLI-Fr, XNLI-Fr, and RTE3-Fr, are based on English datasets that have been machine-translated into French. 
While these provide valuable data for cross-lingual comparison, they may contain artifacts or \enquote{translationese} that do not reflect the full complexity and naturalness of native French. 
Such reliance on translated text for evaluation can be a shortcut that fails to capture the complete challenges of a language, a known issue in cross-lingual transfer methodologies \citep{artetxe-etal-2020-translation}.

\paragraph{Evaluation Scope} Our experimental setup is exclusively focused on a zero-shot evaluation paradigm. 
This approach is valuable for assessing the out-of-the-box capabilities of pretrained models. 
However, it does not evaluate other important aspects of model performance, such as their ability to learn in-context (few-shot learning) or their adaptability through fine-tuning. 
A complete picture of a model's utility requires evaluation across different adaptation strategies, not just a single point of assessment \citep{liang2022holistic}.

\paragraph{Composite Score Granularity} The \name{} composite score is an unweighted arithmetic mean of the individual task scores. 
This straightforward approach ensures simplicity and interpretability but treats all tasks as equally important. 
It does not account for differences in task difficulty, dataset size, or the specific linguistic phenomena being tested. 
Over-reliance on a single aggregate score can be misleading and may fail to highlight the specific strengths and weaknesses of different models, a known pitfall of current benchmarking practices \citep{bowman-dahl-2021-will}.

\section*{Ethical Considerations}
The development and release of the \name{} benchmark, like any tool for advancing language model capabilities, carries ethical implications that warrant careful consideration.

\paragraph{Intended Use and Dual Nature of LLMs}
Our primary goal in creating \name{} is to provide a robust tool for the French NLP community to measure progress and gain a deeper understanding of model capabilities. 
However, we acknowledge that advancements spurred by such benchmarks contribute to the development of more powerful LLMs. 
These models have a dual-use nature: while they can be used for beneficial applications, they can also be exploited for malicious purposes, such as generating convincing disinformation, automating social manipulation, or creating harmful content at scale \citep{bender2021dangers}.

\paragraph{Bias and Representational Harms}
The datasets comprising \name{} are sourced from public domains, including web reviews (Allociné) and Wikipedia articles (FQuAD). 
These sources are known to contain societal biases related to gender, race, religion, and other demographics. 
By using these datasets for evaluation, our benchmark may inadvertently favour models that learn and reproduce these biases. 
We did not perform a comprehensive audit for such biases in the constituent datasets. 
We advocate for users of \name{} to be aware of this and recommend that future work include better documentation and characterization of dataset contents, following principles like those proposed for Datasheets for Datasets \citep{gebru2021datasheets}.

\paragraph{Data Provenance and Privacy}
The data used in \name{}'s datasets, while publicly available, was created by individuals who did not explicitly consent to its use in training or evaluating large-scale AI models. 
For instance, reviews on Allociné were written for other users, not for machine learning research. 
The practice of scraping and repurposing public data without the informed consent of the original creators raises significant ethical questions about privacy and ownership, a problem that has been highlighted in the context of other large-scale web-derived datasets \citep{birhane-etal-2021-multimodal}.

\paragraph{Mitigation and Positive Impact}
Despite the risks, we believe that public benchmarks like \name{} are essential for transparency and accountability in AI. 
By providing a standardized evaluation suite, our work enables researchers to audit proprietary and open models for specific failings, including biases or reasoning gaps. 
Furthermore, \name{} can be used proactively to develop more robust and safer models. 
For example, it can serve as a foundation for \enquote{red teaming} exercises, where the goal is to systematically find and mitigate model harms before deployment \citep{ganguli2022red}. 
We encourage the use of \name{} not only for performance ranking but also for critical safety and ethics research.

\section*{Acknowledgements}
%Removed for double-anonymized.
This research was made possible thanks to the support of a Canadian insurance company, NSERC research grant RDCPJ 537198-18 and FRQNT doctoral research grant. We thank the reviewers for their comments regarding our work.

\bibliography{custom}
\bibliographystyle{acl_natbib}

\appendix
\section{Inference Prompt Details}
\label{ann:prompts}
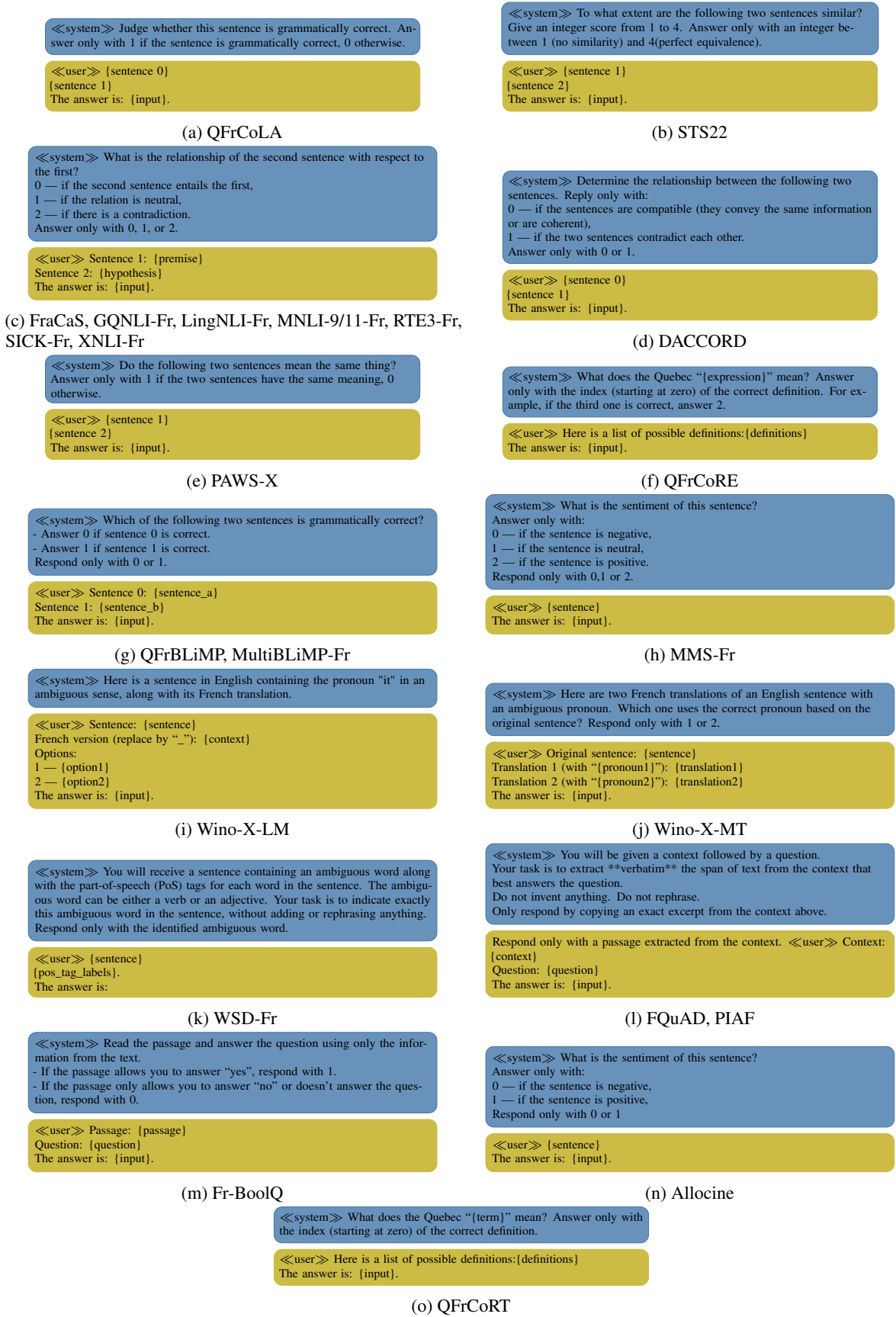
\begin{figure*}[ht!]
    \centering
    \tiny
    \begin{subfigure}{0.49\linewidth}
        \centering
        \begin{tikzpicture}[scale=1, every node/.style={transform shape}]
        \node[rectangle, rounded corners, draw=tgb1, fill=tgb1!80, text width=0.8\linewidth, align=left, inner sep=1ex] (prompt) {
        $\ll$system$\gg$ Judge whether this sentence is grammatically correct. Answer only with 1 if the sentence is grammatically correct, 0 otherwise.
        };
        \node[rectangle, rounded corners, draw=tgb4, fill=tgb4, below=0.1cm of prompt,  text width=0.8\linewidth, align=left, inner sep=1ex] (input) {
        $\ll$user$\gg$ 
        \{sentence 0\}\\
        \{sentence 1\}\\
        The answer is: \{input\}.
        };
        \end{tikzpicture}
        \caption{QFrCoLA}
        \label{fig:qfrcola}
    \end{subfigure}
    \begin{subfigure}{0.49\linewidth}
        \centering
        \begin{tikzpicture}[scale=1, every node/.style={transform shape}]
        \node[rectangle, rounded corners, draw=tgb1, fill=tgb1!80, text width=0.8\linewidth, align=left, inner sep=1.2ex] (prompt) {
        $\ll$system$\gg$ To what extent are the following two sentences similar? Give an integer score from 1 to 4. Answer only with an integer between 1 (no similarity) and 4(perfect equivalence).
        };
        \node[rectangle, rounded corners, draw=tgb4, fill=tgb4, below=0.1cm of prompt,  text width=0.8\linewidth, align=left, inner sep=1ex] (input) {
        $\ll$user$\gg$ 
        \{sentence 1\}\\
        \{sentence 2\}\\
        The answer is: \{input\}.
        };
        \end{tikzpicture}
        \caption{STS22}
        \label{fig:sts22}
    \end{subfigure}
    \begin{subfigure}{0.49\linewidth}
        \centering
        \begin{tikzpicture}[scale=1, every node/.style={transform shape}]
            \node[rectangle, rounded corners, draw=tgb1, fill=tgb1!80, text width=0.87\linewidth, align=left, inner sep=1.2ex] (prompt) {
                $\ll$system$\gg$ What is the relationship of the second sentence with respect to the first?\\
                0 — if the second sentence entails the first,\\
                1 — if the relation is neutral,\\
                2 — if there is a contradiction.\\
                Answer only with 0, 1, or 2.
            };
            \node[rectangle, rounded corners, draw=tgb4, fill=tgb4, below=0.1cm of prompt, text width=0.87\linewidth, align=left, inner sep=1.2ex] (input) {
                $\ll$user$\gg$ Sentence 1: \{premise\}\\Sentence 2: \{hypothesis\}\\The answer is: \{input\}.
            };
        \end{tikzpicture}
        \caption{FraCaS, GQNLI-Fr, LingNLI-Fr, MNLI-9/11-Fr, RTE3-Fr, SICK-Fr, XNLI-Fr}
        \label{fig:gqnli_fr__xnli__mnli_911_fr_mt__sick_fr__rte3_fr}
    \end{subfigure}
    \begin{subfigure}{0.49\linewidth}
        \centering
        \begin{tikzpicture}[scale=1, every node/.style={transform shape}]
        \node[rectangle, rounded corners, draw=tgb1, fill=tgb1!80, text width=0.8\linewidth, align=left, inner sep=1ex] (prompt) {
        $\ll$system$\gg$ Determine the relationship between the following two sentences. Reply only with:\\
        0 — if the sentences are compatible (they convey the same information or are coherent),\\
        1 — if the two sentences contradict each other.\\
        Answer only with 0 or 1.
        };
        \node[rectangle, rounded corners, draw=tgb4, fill=tgb4, below=0.1cm of prompt,  text width=0.8\linewidth, align=left, inner sep=1ex] (input) {
        $\ll$user$\gg$
        \{sentence 0\}\\
        \{sentence 1\}\\
        The answer is: \{input\}.
        };
        \end{tikzpicture}
        \caption{DACCORD}
        \label{fig:daccord}
    \end{subfigure}
    \begin{subfigure}{0.49\linewidth}
        \centering
        \begin{tikzpicture}[scale=1, every node/.style={transform shape}]
        \node[rectangle, rounded corners, draw=tgb1, fill=tgb1!80,text width=0.8\linewidth, align=left, inner sep=1ex] (prompt) {
        $\ll$system$\gg$ Do the following two sentences mean the same thing? Answer only with 1 if the two sentences have the same meaning, 0 otherwise.
        };
        \node[rectangle, rounded corners, draw=tgb4, fill=tgb4, below=0.1cm of prompt,  text width=0.8\linewidth, align=left, inner sep=1ex] (input) {
        $\ll$user$\gg$ 
        \{sentence 1\}\\
        \{sentence 2\}\\
        The answer is: \{input\}.
        };
        \end{tikzpicture}
        \caption{PAWS-X}
        \label{fig:paws_x}
    \end{subfigure}
    \begin{subfigure}{0.49\linewidth}
        \centering
        \begin{tikzpicture}[scale=1, every node/.style={transform shape}]
        \node[rectangle, rounded corners, draw=tgb1, fill=tgb1!80, text width=0.8\linewidth, align=left, inner sep=1ex] (prompt) {
        $\ll$system$\gg$ What does the Quebec \enquote{\{expression\}} mean? 
        Answer only with the index (starting at zero) of the correct definition. For example, if the third one is correct, answer 2.
        };
        \node[rectangle, rounded corners, draw=tgb4, fill=tgb4, below=0.1cm of prompt,  text width=0.8\linewidth, align=left, inner sep=1ex] (input) {
        $\ll$user$\gg$ 
        Here is a list of possible definitions:\{definitions\}\\
        The answer is: \{input\}.
        };
        \end{tikzpicture}
        \caption{QFrCoRE}
        \label{fig:QFrCoRE}
    \end{subfigure}
    \begin{subfigure}{0.49\linewidth}
        \centering
        \begin{tikzpicture}[scale=1, every node/.style={transform shape}]
            \node[rectangle, rounded corners, draw=tgb1, fill=tgb1!80, text width=0.87\linewidth, align=left, inner sep=1.2ex] (prompt) {
                $\ll$system$\gg$ Which of the following two sentences is grammatically correct?\\
                - Answer 0 if sentence 0 is correct.\\
                - Answer 1 if sentence 1 is correct.\\
                Respond only with 0 or 1.
            };
            \node[rectangle, rounded corners, draw=tgb4, fill=tgb4, below=0.1cm of prompt, text width=0.87\linewidth, align=left, inner sep=1.2ex] (input) {
                $\ll$user$\gg$ Sentence 0: \{sentence\_a\}\\Sentence 1: \{sentence\_b\}\\The answer is: \{input\}.
            };
        \end{tikzpicture}
        \caption{QFrBLiMP, MultiBLiMP-Fr}
        \label{fig:qfrblimp}
    \end{subfigure}
    \begin{subfigure}{0.49\linewidth}
        \centering
        \begin{tikzpicture}[scale=1, every node/.style={transform shape}]
            \node[rectangle, rounded corners, draw=tgb1, fill=tgb1!80, text width=0.87\linewidth, align=left, inner sep=1.2ex] (prompt) {
                $\ll$system$\gg$ What is the sentiment of this sentence?\\
                Answer only with:\\
                0 — if the sentence is negative,\\
                1 — if the sentence is neutral,\\
                2 — if the sentence is positive.\\
                 Respond only with 0,1 or 2. 
            };
            \node[rectangle, rounded corners, draw=tgb4, fill=tgb4, below=0.1cm of prompt, text width=0.87\linewidth, align=left, inner sep=1.2ex] (input) {
                $\ll$user$\gg$ \{sentence\}\\
                The answer is: \{input\}.
            };
        \end{tikzpicture}
        \caption{MMS-Fr}
        \label{fig:mms_fr}
    \end{subfigure}
     \begin{subfigure}{0.49\linewidth}
        \centering
        \begin{tikzpicture}[scale=1, every node/.style={transform shape}]
            \node[rectangle, rounded corners, draw=tgb1, fill=tgb1!80, text width=0.87\linewidth, align=left, inner sep=1.2ex] (prompt) {
                $\ll$system$\gg$ Here is a sentence in English containing the pronoun "it" in an ambiguous sense, along with its French translation.

            };
            \node[rectangle, rounded corners, draw=tgb4, fill=tgb4, below=0.1cm of prompt, text width=0.87\linewidth, align=left, inner sep=1.2ex] (input) {
                $\ll$user$\gg$ Sentence: \{sentence\}\\French version (replace by \enquote{\_}): \{context\}\\Options:\\1 — \{option1\}\\2 — \{option2\}\\The answer is: \{input\}.
            };
        \end{tikzpicture}
        \caption{Wino-X-LM}
        \label{fig:lm-winox}
    \end{subfigure}
    \begin{subfigure}{0.49\linewidth}
        \centering
        \begin{tikzpicture}[scale=1, every node/.style={transform shape}]
            \node[rectangle, rounded corners, draw=tgb1, fill=tgb1!80, text width=0.87\linewidth, align=left, inner sep=1.2ex] (prompt) {
                $\ll$system$\gg$ Here are two French translations of an English sentence with an ambiguous pronoun. Which one uses the correct pronoun based on the original sentence? Respond only with 1 or 2.
            };
            \node[rectangle, rounded corners, draw=tgb4, fill=tgb4, below=0.1cm of prompt, text width=0.87\linewidth, align=left, inner sep=1.2ex] (input) {
                $\ll$user$\gg$ Original sentence: \{sentence\}\\
                Translation 1 (with \enquote{\{pronoun1\}}): \{translation1\}\\
                Translation 2 (with \enquote{\{pronoun2\}}): \{translation2\}\\
                The answer is: \{input\}.
            };
        \end{tikzpicture}
        \caption{Wino-X-MT}
        \label{fig:mt-winox}
    \end{subfigure}
    \begin{subfigure}{0.49\linewidth}
        \centering
        \begin{tikzpicture}[scale=1, every node/.style={transform shape}]
            \node[rectangle, rounded corners, draw=tgb1, fill=tgb1!80, text width=0.87\linewidth, align=left, inner sep=1.2ex] (prompt) {
                $\ll$system$\gg$ You will receive a sentence containing an ambiguous word along with the part-of-speech (PoS) tags for each word in the sentence. The ambiguous word can be either a verb or an adjective.
                Your task is to indicate exactly this ambiguous word in the sentence, without adding or rephrasing anything.
                Respond only with the identified ambiguous word.
            };
            \node[rectangle, rounded corners, draw=tgb4, fill=tgb4, below=0.1cm of prompt, text width=0.87\linewidth, align=left, inner sep=1.2ex] (input) {
                $\ll$user$\gg$ \{sentence\}\\
                 \{pos\_tag\_labels\}.\\
                 The answer is:
                 
            };
        \end{tikzpicture}
        \caption{WSD-Fr}
        \label{fig:wsd}
    \end{subfigure}
    \begin{subfigure}{0.49\linewidth}
        \centering
        \begin{tikzpicture}[scale=1, every node/.style={transform shape}]
            \node[rectangle, rounded corners, draw=tgb1, fill=tgb1!80, text width=0.87\linewidth, align=left, inner sep=1.2ex] (prompt) {
                $\ll$system$\gg$ You will be given a context followed by a question.\\
                Your task is to extract **verbatim** the span of text from the context that best answers the question.\\
                Do not invent anything. Do not rephrase.\\
                Only respond by copying an exact excerpt from the context above.\\
            };
            \node[rectangle, rounded corners, draw=tgb4, fill=tgb4, below=0.1cm of prompt, text width=0.87\linewidth, align=left, inner sep=1.2ex] (input) {
                Respond only with a passage extracted from the context. 
                $\ll$user$\gg$ Context: \{context\}\\
                Question: \{question\}\\
                The answer is: \{input\}.
            };
        \end{tikzpicture}
        \caption{FQuAD, PIAF}
        \label{fig:fquad__piaf}
    \end{subfigure}
    \begin{subfigure}{0.49\linewidth}
        \centering
        \begin{tikzpicture}[scale=1, every node/.style={transform shape}]
            \node[rectangle, rounded corners, draw=tgb1, fill=tgb1!80, text width=0.87\linewidth, align=left, inner sep=1.2ex] (prompt) {
                $\ll$system$\gg$ Read the passage and answer the question using only the information from the text.\\
                - If the passage allows you to answer \enquote{yes}, respond with 1.\\
                - If the passage only allows you to answer \enquote{no} or doesn't answer the question, respond with 0.
            };
            \node[rectangle, rounded corners, draw=tgb4, fill=tgb4, below=0.1cm of prompt, text width=0.87\linewidth, align=left, inner sep=1.2ex] (input) {
                $\ll$user$\gg$ Passage: \{passage\}\\
                Question: \{question\}\\
                The answer is: \{input\}.
            };
        \end{tikzpicture}
        \caption{Fr-BoolQ}
        \label{fig:french_boolq}
    \end{subfigure}
    \begin{subfigure}{0.49\linewidth}
        \centering
        \begin{tikzpicture}[scale=1, every node/.style={transform shape}]
            \node[rectangle, rounded corners, draw=tgb1, fill=tgb1!80, text width=0.87\linewidth, align=left, inner sep=1.2ex] (prompt) {
                $\ll$system$\gg$ What is the sentiment of this sentence?\\
                Answer only with:\\
                0 — if the sentence is negative,\\
                1 — if the sentence is positive,\\
                 Respond only with 0 or 1 
            };
            \node[rectangle, rounded corners, draw=tgb4, fill=tgb4, below=0.1cm of prompt, text width=0.87\linewidth, align=left, inner sep=1.2ex] (input) {
                $\ll$user$\gg$ \{sentence\}\\
                The answer is: \{input\}.
            };
        \end{tikzpicture}
        \caption{Allocine}
        \label{fig:allocine}
    \end{subfigure}

    \begin{subfigure}{0.49\linewidth}
        \centering
        \begin{tikzpicture}[scale=1, every node/.style={transform shape}]
        \node[rectangle, rounded corners, draw=tgb1, fill=tgb1!80, text width=0.8\linewidth, align=left, inner sep=1ex] (prompt) {
        $\ll$system$\gg$ What does the Quebec \enquote{\{term\}} mean? Answer only with the index (starting at zero) of the correct definition.
        };
        \node[rectangle, rounded corners, draw=tgb4, fill=tgb4, below=0.1cm of prompt, text width=0.8\linewidth, align=left, inner sep=1ex] (input) {
        $\ll$user$\gg$ 
        Here is a list of possible definitions:\{definitions\}\\
        The answer is: \{input\}.
        };
        \end{tikzpicture}
        \caption{QFrCoRT}
        \label{fig:QFrCoRT}
    \end{subfigure}

    \caption{
    The translated prompt templates used for the zero-shot evaluation of each task in the \name{} benchmark. Each prompt consists of a system message providing the instruction and a user message containing the \texttt{{input}} placeholder for the data instance.
    \hlcbluefonce{Blue} boxes contain the task instructions. \hlcyellowfonce{Yellow} boxes contain the prefix for the model to continue. Texts in \enquote{\texttt{$\ll\gg$}} are role-tags to be fed to the model.}
    \label{fig:prompts}
\end{figure*}

We present in \autoref{fig:prompts}, the translated prompt used to generate the predictions per task.
Prompt were inspired by \citet{aparovich2025belarusianglue} prompts and prompt engineering best practices \citep{marvin2023prompt, ye2024prompt, li2024generation, bjerg2024tips}.

\section{Selected LLM Details}
\label{an:selectedllmdetails}
We present in \autoref{tab:selectedllm} the comprehensive suite of \textbf{open-source} LLMs we could fit on our hardware\footnote{We rely on three NVIDIA RTX 6000 ADA with 49 GB of memory, without memory pooling, thus the maximum size we can fit is around 32B parameters.} (details in \autoref{tab:selectedllm}), detailing their origins and respective sizes, while in \autoref{tab:selectedprivatellm}, we present the comprehensive suite of \textbf{private} LLMs benchmarked in our study.
The selection was curated to cover a wide spectrum of parameter counts, and to include those with specializations in French ($\Upsilon$) or reasoning ($\Gamma$).
All LLMs are downloaded from the \href{https://huggingface.co/models}{HuggingFace Model repository} \citep{wolf2020huggingfaces} using default parameters.

\begin{table*}
    \centering
    \tiny
    \begin{tabular}{llc}
        \toprule
        LLM & Source & Size\\
        \midrule
        \texttt{Apertus-8B-it-2509} & \citet{swissai2025apertus} & 8B \\
        \texttt{Apertus-8B-2509} & \citet{swissai2025apertus} & 8B \\
        \texttt{Aya-23-8b} & \citet{aryabumi2024aya} & 8B \\
        \texttt{Aya-expanse-8b} & \citet{dang2024ayaexpansecombiningresearch} & 8B \\
        \texttt{Chocolatine-14b-it} ($\Upsilon$) & \citet{chocolatine} & 14B \\
        \texttt{Chocolatine-2-14b-it} ($\Upsilon$) & \citet{chocolatinev2} & 14.8B \\
        \texttt{CroissantLLM-Base} ($\Upsilon$) & \citet{faysse2024croissantllm} & 1.3B \\
        \texttt{DeepSeek-R1-distill-Llama-8b} ($\Gamma$) & \citet{deepseekai2025deepseekr1incentivizingreasoningcapability} & 8.03B\\
        \texttt{DeepSeek-R1-distill-Qwen-14b} ($\Gamma$) & \citet{deepseekai2025deepseekr1incentivizingreasoningcapability} & 14.8B\\
        \texttt{DeepSeek-R1-distill-Qwen-32b} ($\Gamma$) & \citet{deepseekai2025deepseekr1incentivizingreasoningcapability} & 32.8B\\
        \texttt{DeepSeek-R1-distill-Qwen-7b} ($\Gamma$) &  \citet{deepseekai2025deepseekr1incentivizingreasoningcapability}& 7.62B\\
        \texttt{DeepSeek-R1-distill-Qwen3-8b} ($\Gamma$) &  \citet{deepseekai2025deepseekr1incentivizingreasoningcapability}& 5.27B\\
        \texttt{Deepthink-reasoning-14b} ($\Gamma$) & \citet{deepthink2} & 14.8B \\
        \texttt{Deepthink-reasoning-7b} ($\Gamma$) & \citet{deepthink1} & 7.62B \\
        \texttt{French-Alpaca-Llama3-8b-it} ($\Upsilon$, $\Gamma$) & \citet{alpaca} & 8.03B \\
        \texttt{Gemma-2-27b-it} ($\Gamma$) & \citet{mesnard2024gemma} & 27.2B   \\
        \texttt{Gemma-2-27b} ($\Gamma$) & \citet{mesnard2024gemma} & 27.2B   \\
        \texttt{Gemma-2-2b-it} ($\Gamma$) & \citet{mesnard2024gemma} & 27.2B\\
        \texttt{Gemma-2-2b} ($\Gamma$)& \citet{mesnard2024gemma} & 2.6B \\
        \texttt{Gemma-2-9b-it} ($\Gamma$) & \citet{mesnard2024gemma} & 9B \\
        \texttt{Gemma-2-9b} ($\Gamma$)& \citet{mesnard2024gemma} & 9.2B \\
        \texttt{GPT-oss-20b} ($\Gamma$)& \citet{gptoss} & 21.5B \\
        \texttt{Granite$3.2$-8B} & \citet{granite2024granite} & 8.17B \\
        \texttt{Granite$3.3$-8B-base} & \citet{granite2024granite} & 8.17B \\
        \texttt{Granite$3.3$-8B-it} & \citet{granite2024granite} & 8.17B \\
        \texttt{Llama-$3.2$-1b-it} ($\Gamma$) & \citet{grattafiori2024llama} & 1.2B \\
        \texttt{Llama-$3.2$-1b} ($\Gamma$)  & \citet{grattafiori2024llama} & 1.2B \\
        \texttt{Llama-$3.2$-3b-it} ($\Gamma$) & \citet{grattafiori2024llama} & 3.21B\\
        \texttt{Llama-$3.2$-3b} ($\Gamma$) & \citet{grattafiori2024llama} & 3.21B \\
        \texttt{Lucie-7b-it-human-data} ($\Upsilon$) & \citet{openllm2025lucie} & 6.71B \\
        \texttt{Lucie-7b-it} ($\Upsilon$) & \citet{openllm2025lucie} & 6.71B\\
        \texttt{Lucie-7b} ($\Upsilon$) & \citet{openllm2025lucie} & 6.71B \\
        \texttt{Meta-Llama-$3.1$-8b-it} ($\Gamma$) & \citet{grattafiori2024llama} & 8B \\
        \texttt{Meta-Llama-$3.1$-8b} ($\Gamma$) & \citet{grattafiori2024llama} & 8B \\
        \texttt{Mixtral-8x7b-it} & \citet{rastogi2025magistral} & 46.7B \\
        \texttt{Mixtral-8x7b} & \citet{rastogi2025magistral} & 46.7B \\
        \texttt{OLMo-2-32B-it} & \citet{olmo20242olmo2furious} & 32.2B \\
        \texttt{OLMo-2-32B} & \citet{olmo20242olmo2furious} & 32.2B \\
        \texttt{OLMo-2-13B-it}& \citet{olmo20242olmo2furious} & 13.7B \\
        \texttt{OLMo-2-13B} & \citet{olmo20242olmo2furious} & 13.7B \\
        \texttt{OLMo-2-7B-it} & \citet{olmo20242olmo2furious} &  7.3B  \\
        \texttt{OLMo-2-7B} & \citet{olmo20242olmo2furious} &  7.3B \\
        \texttt{OLMo-2-1B-it} & \citet{olmo20242olmo2furious} &  1.48B  \\
        \texttt{OLMo-2-1B} & \citet{olmo20242olmo2furious} &  1.48B \\
        \texttt{Phi-$3.5$-mini-it} & \citet{abdin2024phi3technicalreporthighly} & 3.8B \\
        \texttt{Phi-4} & \citet{abdin2024phi} & 14.7B \\
        \texttt{QwQ-32b} ($\Gamma$) & \citet{qwq32b} & 32.8B \\
        \texttt{Qwen$2.5$-$0.5$b-it} ($\Gamma$) & \citet{hui2024qwen2}  & 494M \\
        \texttt{Qwen$2.5$-$0.5$b} & \citet{hui2024qwen2} & 494M \\
        \texttt{Qwen$2.5$-$1.5$b-it} & \citet{hui2024qwen2}  & 1.5B \\
        \texttt{Qwen$2.5$-$1.5$b} & \citet{hui2024qwen2}  & 1.5B \\
        \texttt{Qwen$2.5$-14b-it} & \citet{hui2024qwen2}  & 14.7B \\
        \texttt{Qwen$2.5$-14b} & \citet{hui2024qwen2}  & 14.7B \\
        \texttt{Qwen$2.5$-32b-it} & \citet{hui2024qwen2}  & 32.8B \\
        \texttt{Qwen$2.5$-32b} & \citet{hui2024qwen2}  & 32.8B \\
        \texttt{Qwen$2.5$-3b-it} & \citet{hui2024qwen2}  & 3B \\
        \texttt{Qwen$2.5$-3b} & \citet{hui2024qwen2}  & 3B \\
        \texttt{Qwen$2.5$-7b-it} & \citet{hui2024qwen2}  & 7.6B \\
        \texttt{Qwen$2.5$-7b} & \citet{hui2024qwen2}  & 7.6B \\
        \texttt{Qwen3-14b-base} & \citet{qwen3technicalreport}  & 14.8B \\
        \texttt{Qwen3-14b} & \citet{qwen3technicalreport}  & 8.76B \\
        \texttt{Reka-flash-3} ($\Gamma$) & \citet{reka} & 20.9B \\
        \texttt{S1.1-32b} ($\Gamma$) & \citet{s11} & 32.8B \\
        \texttt{SmolLM2-$1.7$b-it} & \citet{allal2025smollm2smolgoesbig}  & 1.7B \\
        \texttt{SmolLM2-$1.7$b} & \citet{allal2025smollm2smolgoesbig}  & 1.7B \\
        \texttt{SmolLM2-135m-it} & \citet{allal2025smollm2smolgoesbig}  & 134.5M \\  
        \texttt{SmolLM2-135m} &\citet{allal2025smollm2smolgoesbig}  & 134.5M \\
        \texttt{SmolLM2-360m-it} & \citet{allal2025smollm2smolgoesbig}  & 361.8M \\
        \texttt{SmolLM2-360m} & \citet{allal2025smollm2smolgoesbig}  & 361.8M \\
        \bottomrule
    \end{tabular}%
    \caption{The selected open-source LLM used in our work, along with their source and size. \enquote{$\Upsilon$} are model that have a specialization in French, while \enquote{$\Gamma$} are model marketed as reasoning LLM.}
    \label{tab:selectedllm}
\end{table*}

\begin{table*}
    \centering
    \begin{tabular}{ll}
        \toprule
        LLM & Source\\
        \midrule
        \texttt{Claude-Opus-4-20250514} ($\Gamma$) & Anthropic\\
        \texttt{Claude-Sonnet-4-20250514} ($\Gamma$) & Anthropic\\
        \texttt{DeepSeek-chat} & DeepSeek\\
        \texttt{DeepSeek-reasoner}  ($\Gamma$) & DeepSeek\\
        \texttt{Gemini-$2.5$-flash} & Google \\
        \texttt{Gemini-$2.5$-pro} ($\Gamma$) & Google \\
        \texttt{GPT-$4.1$-2025-04-14} &OpenAI \\
        \texttt{GPT-$4.1$-mini-2025-04-14} &OpenAI \\
        \texttt{GPT-4o-2024-08-06} &OpenAI \\
        \texttt{GPT-4o-mini-2024-07-18} &OpenAI \\
        \texttt{GPT-5-2025-08-07} ($\Gamma$)& OpenAI\\
        \texttt{GPT-5-mini-2025-08-07} ($\Gamma$)& OpenAI\\
        \texttt{GPT-oss-120B} ($\Gamma$)& OpenAI\\
        \texttt{Grok-3-fast-latest} ($\Gamma$) & xAI \\
        \texttt{Grok-3-latest} ($\Gamma$)& xAI\\
        \texttt{Grok-3-mini-fast-latest} ($\Gamma$)& xAI\\
        \texttt{Grok-3-mini-latest} ($\Gamma$)& xAI\\
        \texttt{Grok-4-0709} ($\Gamma$) & xAI \\
        \texttt{Mistral-large-latest} ($\Gamma$)& Mistral\\
        \texttt{o1-2024-12-17} ($\Gamma$)&OpenAI\\
        \texttt{o1-mini-2024-09-12} ($\Gamma$)&OpenAI\\
        \texttt{o3-2025-04-16} ($\Gamma$)&OpenAI\\
        \texttt{o3-mini-2025-01-31} ($\Gamma$)&OpenAI\\
        \texttt{Pixtral-large-latest} & Mistral\\
        \texttt{Qwen-max} ($\Gamma$) & Qwen \\
        \bottomrule
    \end{tabular}%
    \caption{The selected private LLM used in our work, along with their source. \enquote{$\Gamma$} are model that are marketed as reasoning LLM.}
    \label{tab:selectedprivatellm}
\end{table*}

\section{Hardware and Private LLM Inference Budget}
\label{an:hardware}
\subsection{Hardware}
We rely on three NVIDIA RTX 6000 ADA with 49 GB of memory, without memory pooling; thus, the maximum size we can fit is around 32B parameters in order to have a sufficient batch size to process the experiment in a reasonable timeframe (i.e. a month or so).

\subsection{Private LLM Inference Budget}
We allocated a budget of approximately 2,000 USD for using private LLM APIs (e.g., OpenAI, Anthropic) during development, prototyping, and adjusting our prompts. For the complete inference loop across all selected private LLMs and tasks, we allocated a budget of nearly \$ 17,500 USD.
It took approximately four weeks to process all private model inference in parallel.

\section{Complete Experiments Results}
\label{an:results_complete}
In this section, we present our complete results in \autoref{tab:results_complete}.

\begin{table*}[ht!]
    \centering
    \resizebox{\textwidth}{!}{%
    \begin{tabular}{lRRRRRRRRRRRRRRRRRRRRRRRRR}
        \toprule
        LLM &  \multicolumn{1}{c}{\begin{tabular}[c]{@{}c@{}}Allocine\\ Acc. (\%)\end{tabular}} & 
        \multicolumn{1}{c}{\begin{tabular}[c]{@{}c@{}}DACCORD\\ Acc. (\%)\end{tabular}} & 
        \multicolumn{1}{c}{\begin{tabular}[c]{@{}c@{}}FQuAD\\EM (\%)\end{tabular}} & 
        \multicolumn{1}{c}{\begin{tabular}[c]{@{}c@{}}FQuAD\\F1 (\%)\end{tabular}} & 
        \multicolumn{1}{c}{\begin{tabular}[c]{@{}c@{}}FraCaS\\ Acc. (\%)\end{tabular}} & 
        \multicolumn{1}{c}{\begin{tabular}[c]{@{}c@{}}Fr-BoolQ\\ Acc. (\%)\end{tabular}} & 
        \multicolumn{1}{c}{\begin{tabular}[c]{@{}c@{}}GQNLI-Fr\\ Acc. (\%)\end{tabular}} & 
        \multicolumn{1}{c}{\begin{tabular}[c]{@{}c@{}}LingNLI-Fr\\ Acc. (\%)\end{tabular}} &
        \multicolumn{1}{c}{\begin{tabular}[c]{@{}c@{}}MMS\\ Acc. (\%)\end{tabular}} & 
        \multicolumn{1}{c}{\begin{tabular}[c]{@{}c@{}}MNLI-9/11-Fr\\ Acc. (\%)\end{tabular}} & 
        \multicolumn{1}{c}{\begin{tabular}[c]{@{}c@{}}MultiBLiMP-Fr\\ Acc. (\%)\end{tabular}} & 
        \multicolumn{1}{c}{\begin{tabular}[c]{@{}c@{}}PAWS-X\\ Acc. (\%)\end{tabular}} & 
        \multicolumn{1}{c}{\begin{tabular}[c]{@{}c@{}}PIAF\\ EM (\%)\end{tabular}} & 
        \multicolumn{1}{c}{\begin{tabular}[c]{@{}c@{}}PIAF\\F1 (\%)\end{tabular}} & 
        \multicolumn{1}{c}{\begin{tabular}[c]{@{}c@{}}QFrBLiMP\\ Acc. (\%)\end{tabular}} & 
        \multicolumn{1}{c}{\begin{tabular}[c]{@{}c@{}}QFrCoLA\\ Acc. (\%)\end{tabular}} & 
        \multicolumn{1}{c}{\begin{tabular}[c]{@{}c@{}}QFrCoRE\\ Acc. (\%)\end{tabular}} & 
        \multicolumn{1}{c}{\begin{tabular}[c]{@{}c@{}}QFrCoRT\\ Acc. (\%)\end{tabular}} & 
        \multicolumn{1}{c}{\begin{tabular}[c]{@{}c@{}}RT3-Fr\\ Acc. (\%)\end{tabular}} & 
        \multicolumn{1}{c}{\begin{tabular}[c]{@{}c@{}}SICK-Fr\\ Acc. (\%)\end{tabular}} & 
        \multicolumn{1}{c}{\begin{tabular}[c]{@{}c@{}}STS22\\ Acc. (\%)\end{tabular}} & 
        \multicolumn{1}{c}{\begin{tabular}[c]{@{}c@{}}Wino-X-LM\\ Acc. (\%)\end{tabular}} & 
        \multicolumn{1}{c}{\begin{tabular}[c]{@{}c@{}}Wino-X-MT\\ Acc. (\%)\end{tabular}} & 
        \multicolumn{1}{c}{\begin{tabular}[c]{@{}c@{}}WSD\\ EM (\%)\end{tabular}} & 
        \multicolumn{1}{c}{\begin{tabular}[c]{@{}c@{}}XNLI-Fr\\ Acc. (\%)\end{tabular}} \\
\midrule
\texttt{Apertus-8B-2509} & 48.83 & 49.61 & 25.00 & 6.37 & 14.03 & 46.07 & 26.67 & 31.70 & 20.97 & 31.70 & 40.26 & 49.55 & 26.04 & 5.73 & 50.28 & 42.30 & 9.76 & 14.04 & 38.12 & 49.86 & 19.44 & 49.98 & 49.70 & 0.00 & 33.41 \\
\texttt{Apertus-8B-it-2509} & 51.59 & 50.19 & 50.00 & 18.19 & 11.94 & 49.44 & 30.00 & 32.97 & 38.76 & 33.65 & 49.35 & 54.90 & 0.00 & 14.32 & 51.80 & 31.61 & 13.40 & 18.13 & 10.12 & 31.31 & 22.22 & 50.66 & 49.53 & 0.00 & 32.65 \\
\texttt{Aya-23-8b} & 46.20 & 49.90 & 25.00 & 21.05 & 9.85 & 48.88 & 30.00 & 32.03 & 39.48 & 32.70 & 45.45 & 48.55 & 1.04 & 17.53 & 50.47 & 62.67 & 9.99 & 12.87 & 9.88 & 15.37 & 26.39 & 49.12 & 49.93 & 0.00 & 33.33 \\
\texttt{Aya-expanse-8b} & 48.58 & 49.81 & 27.75 & 64.38 & 29.25 & 46.07 & 30.00 & 32.56 & 20.53 & 32.05 & 45.45 & 48.10 & 20.05 & 54.52 & 52.93 & 67.92 & 9.37 & 7.60 & 39.75 & 56.87 & 29.17 & 49.19 & 50.64 & 0.83 & 33.33 \\
\texttt{Chocolatine-14B-it-DPO-v$1.3$} ($\Upsilon$) & 53.05 & 60.83 & 50.00 & 4.41 & 28.06 & 50.00 & 36.67 & 33.58 & 33.95 & 33.40 & 59.74 & 54.90 & 0.00 & 0.00 & 64.27 & 30.93 & 9.61 & 10.53 & 28.75 & 31.72 & 29.17 & 57.72 & 49.97 & 2.92 & 34.55 \\
\texttt{Chocolatine-2-14B-it-v$2.0$.3} ($\Upsilon$) & 80.73 & 54.84 & 21.50 & 55.57 & 47.46 & 44.38 & 36.67 & 36.36 & 52.47 & 36.35 & 94.81 & 46.10 & 17.19 & 46.52 & 86.01 & 72.52 & 11.33 & 12.28 & 45.12 & 23.93 & 30.56 & 64.63 & 53.08 & 17.85 & 37.98 \\
\texttt{Claude-opus-4-20250514} ($\Gamma$) & 96.92 & 95.07 & 0.00 & 0.00 & 61.79 & 94.94 & 60.00 & 69.55 & 75.65 & 70.30 & 98.70 & 79.05 & 26.04 & 29.11 & 88.66 & 82.31 & 93.46 & 97.66 & 82.88 & 75.64 & 51.39 & 91.34 & 82.13 & 0.16 & 75.63 \\
\texttt{Claude-sonnet-4-20250514} ($\Gamma$) & 96.73 & 96.62 & 0.00 & 2.78 & 58.21 & 96.63 & 63.33 & 72.70 & 75.29 & 72.70 & 97.40 & 75.25 & 0.00 & 10.42 & 88.85 & 82.63 & 91.75 & 97.66 & 86.75 & 78.88 & 58.33 & 92.23 & 73.59 & 1.51 & 73.69 \\
\texttt{CroissantLLMBase} ($\Upsilon$) & 47.98 & 49.81 & 0.00 & 2.18 & 9.85 & 48.31 & 30.00 & 32.05 & 39.50 & 32.70 & 45.45 & 48.55 & 0.00 & 59.80 & 49.53 & 59.91 & 10.58 & 9.36 & 9.12 & 14.51 & 22.22 & 51.13 & 49.43 & 0.00 & 33.33 \\
\texttt{DeepSeek-R1-0528-Qwen3-8B} & 30.18 & 47.10 & 0.00 & 7.80 & 37.01 & 50.00 & 40.00 & 33.78 & 12.11 & 33.65 & 53.25 & 49.80 & 0.00 & 7.41 & 51.23 & 48.66 & 3.93 & 6.43 & 29.88 & 45.64 & 13.89 & 48.84 & 50.10 & 0.00 & 25.79 \\
\texttt{DeepSeek-R1-Distill-Llama-8B} ($\Gamma$) & 57.60 & 52.80 & 0.00 & 6.93 & 20.60 & 43.82 & 46.67 & 33.89 & 42.70 & 32.60 & 50.65 & 48.10 & 0.00 & 8.20 & 51.23 & 55.96 & 10.49 & 11.11 & 41.38 & 15.39 & 26.39 & 49.27 & 49.03 & 0.00 & 32.87 \\
\texttt{DeepSeek-R1-Distill-Qwen-14B} ($\Gamma$) & 68.64 & 72.53 & 0.00 & 5.05 & 44.18 & 46.63 & 40.00 & 43.18 & 71.36 & 45.45 & 41.56 & 48.70 & 0.00 & 3.74 & 50.09 & 62.70 & 32.18 & 35.67 & 31.87 & 20.77 & 26.39 & 50.59 & 49.46 & 0.00 & 53.55 \\
\texttt{DeepSeek-R1-Distill-Qwen-32B} ($\Gamma$) & 94.21 & 58.99 & 0.00 & 8.58 & 51.04 & 65.17 & 36.67 & 47.70 & 73.17 & 51.40 & 58.44 & 54.50 & 0.00 & 7.94 & 60.11 & 67.45 & 31.04 & 53.22 & 54.50 & 55.01 & 38.89 & 56.25 & 50.03 & 0.00 & 48.70 \\
\texttt{DeepSeek-R1-Distill-Qwen-7B} ($\Gamma$) & 41.05 & 50.87 & 0.00 & 4.28 & 48.36 & 52.81 & 23.33 & 33.50 & 33.75 & 34.25 & 49.35 & 50.25 & 0.00 & 6.83 & 52.93 & 64.31 & 8.72 & 12.87 & 50.62 & 56.67 & 27.78 & 50.38 & 48.09 & 0.00 & 32.61 \\
\texttt{DeepSeek-chat} & 95.20 & 92.75 & 37.00 & 63.48 & 35.22 & 92.70 & 26.67 & 61.27 & 73.11 & 63.50 & 97.40 & 75.80 & 24.74 & 49.75 & 88.85 & 80.81 & 83.92 & 92.40 & 70.25 & 33.49 & 50.00 & 79.45 & 52.91 & 44.79 & 64.61 \\
\texttt{DeepSeek-reasoner} ($\Gamma$) & 95.71 & 96.23 & 7.75 & 13.54 & 63.88 & 93.26 & 46.67 & 62.17 & 73.90 & 63.60 & 97.40 & 77.45 & 7.81 & 14.39 & 89.98 & 81.08 & 84.98 & 91.23 & 74.25 & 69.71 & 52.78 & 77.77 & 53.08 & 0.26 & 66.05 \\
\texttt{Deepthink-Reasoning-14B} ($\Gamma$) & 29.39 & 70.50 & 1.25 & 26.18 & 29.55 & 53.93 & 33.33 & 42.20 & 40.43 & 50.45 & 89.61 & 53.75 & 0.00 & 23.50 & 69.94 & 54.90 & 32.87 & 31.58 & 38.50 & 48.12 & 25.00 & 54.74 & 50.20 & 0.00 & 51.92 \\
\texttt{Deepthink-Reasoning-7B} ($\Gamma$) & 55.22 & 50.48 & 1.50 & 4.27 & 42.69 & 55.62 & 53.33 & 36.85 & 70.49 & 41.00 & 49.35 & 54.90 & 52.08 & 5.91 & 52.74 & 65.33 & 15.73 & 13.45 & 60.62 & 36.36 & 37.50 & 52.70 & 51.17 & 0.38 & 39.20 \\
\texttt{French-Alpaca-Llama3-8B-it-v$1.0$} ($\Gamma$$\Upsilon$) & 78.44 & 56.67 & 100.00 & 8.46 & 60.30 & 51.69 & 40.00 & 35.23 & 43.98 & 35.30 & 58.44 & 47.65 & 78.12 & 9.51 & 53.88 & 65.35 & 23.74 & 16.37 & 50.88 & 28.62 & 31.94 & 51.31 & 50.70 & 0.03 & 33.83 \\
\texttt{GPT-$4.1$-2025-04-14} & 96.20 & 96.81 & 0.00 & 0.00 & 63.88 & 95.51 & 70.00 & 73.19 & 75.56 & 75.20 & 98.70 & 77.05 & 0.00 & 0.00 & 89.79 & 82.89 & 86.73 & 94.15 & 83.00 & 80.17 & 50.00 & 84.71 & 62.95 & 1.22 & 71.88 \\
\texttt{GPT-$4.1$-mini-2025-04-14} & 95.43 & 93.91 & 16.00 & 49.07 & 54.93 & 93.82 & 70.00 & 65.77 & 75.18 & 66.90 & 93.51 & 73.45 & 13.80 & 39.00 & 89.22 & 78.98 & 82.80 & 92.40 & 81.12 & 72.85 & 51.39 & 79.38 & 57.76 & 34.57 & 70.68 \\
\texttt{GPT-4o-2024-08-06} & 94.44 & 96.62 & 0.00 & 2.78 & 58.21 & 94.38 & 66.67 & 70.75 & 74.92 & 72.40 & 98.70 & 78.00 & 0.00 & 0.00 & 88.47 & 82.85 & 83.94 & 93.57 & 79.00 & 80.00 & 47.22 & 83.03 & 58.17 & 0.00 & 71.98 \\
\texttt{GPT-4o-mini-2024-07-18} & 95.52 & 93.13 & 6.75 & 43.41 & 66.27 & 97.19 & 53.33 & 63.44 & 74.00 & 63.80 & 97.40 & 77.10 & 5.99 & 33.87 & 86.77 & 81.50 & 73.88 & 90.64 & 76.62 & 77.44 & 56.94 & 69.14 & 51.37 & 39.86 & 67.56 \\
\texttt{GPT-5-2025-08-07} ($\Gamma$) & 97.08 & 96.13 & 0.00 & 0.00 & 43.88 & 94.38 & 60.00 & 57.49 & 75.27 & 61.25 & 98.70 & 78.45 & 0.00 & 10.42 & 90.93 & 83.99 & 87.93 & 95.32 & 82.50 & 80.41 & 47.22 & 94.63 & 91.63 & 0.00 & 62.73 \\
\texttt{GPT-5-mini-2025-08-07} ($\Gamma$) & 96.47 & 96.42 & 12.50 & 43.26 & 58.21 & 96.07 & 66.67 & 63.64 & 75.21 & 66.50 & 98.70 & 77.45 & 7.29 & 31.87 & 89.22 & 81.90 & 77.36 & 92.40 & 85.88 & 80.84 & 58.33 & 93.84 & 91.03 & 37.62 & 74.25 \\
\texttt{GPT-oss-120b} ($\Gamma$) & 94.71 & 95.84 & 3.25 & 15.46 & 41.49 & 93.26 & 56.67 & 61.70 & 71.72 & 65.25 & 100.00 & 77.40 & 2.34 & 13.42 & 89.41 & 78.15 & 66.52 & 76.61 & 83.88 & 84.08 & 52.78 & 81.10 & 74.26 & 4.20 & 65.67 \\
\texttt{GPT-oss-20b} ($\Gamma$) & 49.95 & 50.39 & 0.00 & 16.05 & 25.37 & 52.25 & 30.00 & 33.89 & 21.05 & 34.00 & 38.96 & 46.95 & 0.00 & 12.76 & 51.98 & 53.46 & 6.65 & 8.77 & 15.50 & 37.89 & 25.00 & 51.16 & 49.56 & 0.00 & 31.30 \\
\texttt{Gemini-$2.5$-flash} & 89.58 & 95.16 & 0.00 & 0.00 & 60.30 & 95.51 & 63.33 & 66.40 & 72.81 & 66.25 & 100.00 & 77.45 & 0.00 & 0.00 & 88.66 & 80.86 & 87.80 & 95.91 & 79.50 & 72.85 & 48.61 & 82.96 & 61.38 & 0.00 & 69.84 \\
\texttt{Gemini-$2.5$-pro} ($\Gamma$) & 96.58 & 97.39 & 0.00 & 2.78 & 55.82 & 95.51 & 70.00 & 67.89 & 75.50 & 67.50 & 97.40 & 76.95 & 0.00 & 10.42 & 90.36 & 85.77 & 90.68 & 96.49 & 85.62 & 71.40 & 50.00 & 92.52 & 88.42 & 0.00 & 70.68 \\
\texttt{Gemma-2-27b-it} ($\Gamma$) & 17.81 & 50.19 & 3.00 & 4.48 & 54.63 & 53.37 & 40.00 & 32.21 & 27.54 & 30.60 & 59.74 & 50.95 & 2.60 & 3.87 & 58.98 & 44.12 & 27.78 & 15.20 & 45.00 & 35.30 & 25.00 & 55.14 & 49.80 & 1.25 & 31.82 \\
\texttt{Gemma-2-27b} ($\Gamma$) & 44.84 & 49.23 & 75.00 & 2.26 & 31.94 & 51.12 & 33.33 & 31.19 & 21.92 & 30.45 & 40.26 & 50.80 & 78.12 & 1.58 & 46.12 & 48.17 & 6.15 & 4.68 & 21.50 & 28.54 & 27.78 & 48.37 & 50.27 & 0.19 & 30.84 \\
\texttt{Gemma-2-2b-it} ($\Gamma$) & 11.37 & 49.61 & 75.00 & 2.08 & 31.64 & 56.74 & 30.00 & 30.29 & 16.68 & 29.80 & 44.16 & 50.30 & 0.00 & 55.22 & 51.61 & 55.76 & 12.84 & 5.26 & 33.50 & 56.44 & 23.61 & 49.77 & 49.10 & 0.00 & 25.47 \\
\texttt{Gemma-2-2b} ($\Gamma$) & 48.39 & 49.81 & 1.75 & 3.66 & 33.73 & 49.44 & 33.33 & 33.54 & 21.78 & 33.25 & 45.45 & 46.70 & 78.12 & 1.39 & 51.42 & 51.83 & 10.64 & 12.87 & 33.50 & 55.52 & 26.39 & 50.98 & 49.16 & 0.00 & 32.83 \\
\texttt{Gemma-2-9b-it} ($\Gamma$) & 46.59 & 21.08 & 2.50 & 5.92 & 45.07 & 57.87 & 23.33 & 25.04 & 6.01 & 22.85 & 64.94 & 58.55 & 52.08 & 1.15 & 69.19 & 69.32 & 6.76 & 8.19 & 30.63 & 7.32 & 9.72 & 52.85 & 51.54 & 0.42 & 19.54 \\
\texttt{Gemma-2-9b} ($\Gamma$) & 47.77 & 49.71 & 2.25 & 3.81 & 25.97 & 47.19 & 36.67 & 32.60 & 21.50 & 29.40 & 51.95 & 45.80 & 0.00 & 70.85 & 52.55 & 58.73 & 2.18 & 5.85 & 40.38 & 56.22 & 31.94 & 49.98 & 49.90 & 0.03 & 33.27 \\
\texttt{Granite-$3.2$-8b-it} & 92.90 & 49.90 & 14.75 & 48.03 & 60.90 & 51.69 & 40.00 & 35.40 & 48.70 & 35.25 & 45.45 & 49.90 & 12.76 & 41.76 & 54.25 & 67.19 & 11.33 & 10.53 & 51.25 & 28.64 & 30.56 & 50.63 & 51.31 & 16.92 & 33.25 \\
\texttt{Granite-$3.3$-8b-base} & 52.46 & 52.61 & 75.00 & 17.64 & 61.19 & 50.56 & 46.67 & 35.21 & 40.80 & 35.55 & 53.25 & 48.05 & 1.82 & 14.71 & 53.12 & 49.23 & 9.82 & 12.87 & 48.38 & 30.78 & 37.50 & 49.41 & 50.23 & 0.10 & 33.49 \\
\texttt{Granite-$3.3$-8b-it} & 74.72 & 49.32 & 0.00 & 27.02 & 37.01 & 52.25 & 26.67 & 33.21 & 37.58 & 33.70 & 49.35 & 46.65 & 0.00 & 21.45 & 50.85 & 64.90 & 12.67 & 17.54 & 35.38 & 37.22 & 27.78 & 50.66 & 50.33 & 0.00 & 33.61 \\
\texttt{Grok-3-fast-latest} ($\Gamma$) & 95.80 & 96.62 & 0.00 & 0.00 & 57.31 & 95.51 & 60.00 & 67.67 & 74.38 & 70.45 & 98.70 & 77.75 & 0.00 & 0.00 & 89.79 & 82.71 & 79.43 & 92.40 & 82.38 & 73.85 & 41.67 & 83.64 & 61.11 & 0.00 & 67.56 \\
\texttt{Grok-3-latest} ($\Gamma$) & 95.83 & 96.62 & 0.00 & 0.00 & 58.81 & 95.51 & 66.67 & 67.28 & 74.39 & 70.75 & 98.70 & 77.80 & 0.00 & 0.00 & 90.74 & 82.48 & 79.47 & 91.23 & 82.12 & 73.77 & 44.44 & 83.71 & 60.94 & 0.00 & 67.37 \\
\texttt{Grok-3-mini-fast-latest} ($\Gamma$) & 95.87 & 97.49 & 0.00 & 2.78 & 34.03 & 94.94 & 50.00 & 58.19 & 75.35 & 57.70 & 98.70 & 75.80 & 0.00 & 10.42 & 90.36 & 81.74 & 82.54 & 88.89 & 82.50 & 77.48 & 55.56 & 86.25 & 81.83 & 0.00 & 60.84 \\
\texttt{Grok-3-mini-latest} ($\Gamma$) & 95.83 & 97.29 & 0.00 & 2.78 & 35.82 & 94.38 & 56.67 & 57.92 & 75.26 & 58.00 & 100.00 & 76.60 & 0.00 & 10.42 & 90.36 & 82.00 & 82.50 & 90.64 & 82.75 & 78.54 & 54.17 & 86.32 & 81.76 & 0.00 & 60.28 \\
\texttt{Grok-4-0709} & 96.98 & 91.97 & 0.00 & 0.00 & 36.42 & 94.38 & 56.67 & 55.69 & 76.24 & 57.95 & 100.00 & 73.95 & 0.00 & 10.42 & 90.93 & 83.21 & 85.84 & 95.32 & 83.88 & 75.30 & 44.44 & 95.99 & 93.74 & 0.00 & 58.34 \\
\texttt{Llama-$3.2$-1B-it} ($\Gamma$) & 58.78 & 49.90 & 3.00 & 3.96 & 54.93 & 47.75 & 16.67 & 33.17 & 34.03 & 33.15 & 44.16 & 52.50 & 1.04 & 1.94 & 49.34 & 48.30 & 15.93 & 12.28 & 47.25 & 28.70 & 31.94 & 51.63 & 50.33 & 0.03 & 32.97 \\
\texttt{Llama-$3.2$-1B} ($\Gamma$) & 51.93 & 49.52 & 25.00 & 1.43 & 61.49 & 55.62 & 40.00 & 34.91 & 39.98 & 35.25 & 51.95 & 50.70 & 26.04 & 72.41 & 50.09 & 34.75 & 9.67 & 11.11 & 51.25 & 28.62 & 26.39 & 49.91 & 49.90 & 0.00 & 32.51 \\
\texttt{Llama-$3.2$-3B-it} ($\Gamma$) & 54.97 & 59.19 & 3.25 & 6.89 & 29.85 & 60.67 & 43.33 & 38.71 & 70.60 & 37.50 & 51.95 & 50.45 & 26.04 & 3.16 & 55.95 & 60.15 & 17.61 & 24.56 & 49.88 & 41.19 & 30.56 & 49.70 & 49.67 & 0.03 & 40.90 \\
\texttt{Llama-$3.2$-3B} ($\Gamma$) & 51.89 & 50.00 & 75.00 & 2.83 & 13.13 & 50.56 & 33.33 & 31.31 & 39.25 & 32.05 & 46.75 & 54.00 & 0.00 & 53.34 & 48.58 & 31.83 & 9.67 & 12.87 & 12.38 & 14.51 & 33.33 & 50.48 & 51.27 & 0.00 & 33.11 \\
\texttt{Lucie-7B-it-human-data} ($\Upsilon$) & 52.11 & 50.58 & 29.75 & 50.31 & 42.99 & 48.88 & 40.00 & 34.85 & 36.59 & 36.65 & 50.65 & 54.75 & 22.40 & 43.09 & 50.09 & 32.10 & 9.91 & 10.53 & 12.50 & 19.81 & 23.61 & 49.87 & 49.73 & 0.06 & 33.51 \\
\texttt{Lucie-7B-it-v$1.1$} ($\Upsilon$) & 43.36 & 48.74 & 0.00 & 23.69 & 10.45 & 43.82 & 30.00 & 31.76 & 54.80 & 32.85 & 46.75 & 49.95 & 0.00 & 18.87 & 47.83 & 39.11 & 17.35 & 17.54 & 9.00 & 14.51 & 19.44 & 48.48 & 49.73 & 0.00 & 31.58 \\
\texttt{Lucie-7B} ($\Upsilon$) & 47.57 & 49.90 & 35.50 & 55.66 & 60.30 & 46.63 & 36.67 & 35.34 & 39.95 & 35.25 & 51.95 & 54.10 & 30.21 & 47.56 & 51.98 & 31.26 & 10.12 & 7.60 & 42.75 & 19.26 & 20.83 & 49.70 & 49.73 & 2.76 & 33.37 \\
\texttt{Meta-Llama-$3.1$-8B-it} ($\Gamma$) & 82.30 & 74.37 & 4.75 & 8.33 & 40.90 & 55.62 & 36.67 & 38.22 & 56.18 & 39.95 & 62.34 & 54.85 & 3.39 & 11.56 & 62.57 & 65.54 & 20.14 & 9.36 & 41.75 & 58.56 & 37.50 & 50.91 & 48.96 & 0.13 & 38.38 \\
\texttt{Meta-Llama-$3.1$-8B} ($\Gamma$) & 52.48 & 50.10 & 2.25 & 5.65 & 60.90 & 48.88 & 40.00 & 35.32 & 48.26 & 35.20 & 50.65 & 52.35 & 0.00 & 2.75 & 47.64 & 48.22 & 9.78 & 9.36 & 46.25 & 21.83 & 27.78 & 49.87 & 49.33 & 0.00 & 33.25 \\
\texttt{Mistral-large-latest} ($\Gamma$) & 95.76 & 95.07 & 0.00 & 0.00 & 58.81 & 94.38 & 46.67 & 65.89 & 75.31 & 66.15 & 96.10 & 77.50 & 0.00 & 0.00 & 90.17 & 82.18 & 84.03 & 90.64 & 75.62 & 70.83 & 48.61 & 81.38 & 55.89 & 0.00 & 66.51 \\
\texttt{Mixtral-8x7B-it-v$0.1$} & 60.52 & 49.42 & 4.25 & 39.73 & 40.30 & 49.44 & 46.67 & 34.42 & 40.26 & 34.05 & 41.56 & 53.70 & 4.95 & 33.12 & 39.32 & 35.56 & 9.86 & 11.70 & 32.88 & 20.57 & 18.06 & 52.81 & 49.40 & 8.88 & 32.81 \\
\texttt{Mixtral-8x7B-v$0.1$} & 52.06 & 49.71 & 10.50 & 32.99 & 60.30 & 49.44 & 40.00 & 35.07 & 39.88 & 35.50 & 53.25 & 47.50 & 9.11 & 26.52 & 57.84 & 63.99 & 9.43 & 11.70 & 49.38 & 28.01 & 27.78 & 50.73 & 49.10 & 2.76 & 33.37 \\
\texttt{OLMo-2-0325-32B-it} & 3.89 & 47.10 & 13.50 & 50.39 & 40.90 & 37.08 & 16.67 & 31.23 & 15.25 & 31.05 & 36.36 & 43.70 & 11.46 & 41.33 & 40.83 & 47.40 & 3.95 & 2.92 & 27.62 & 23.81 & 26.39 & 50.52 & 50.47 & 3.49 & 13.85 \\
\texttt{OLMo-2-0325-32B} & 46.92 & 50.19 & 11.25 & 37.14 & 58.21 & 42.70 & 36.67 & 31.86 & 19.36 & 34.05 & 46.75 & 47.60 & 7.29 & 28.97 & 43.10 & 45.49 & 8.29 & 5.85 & 27.25 & 32.25 & 33.33 & 49.41 & 48.49 & 1.06 & 32.85 \\
\texttt{OLMo-2-0425-1B-it} & 48.58 & 49.13 & 8.75 & 31.55 & 16.72 & 51.69 & 33.33 & 31.70 & 27.58 & 34.10 & 55.84 & 51.40 & 2.08 & 21.42 & 49.72 & 53.71 & 12.82 & 32.16 & 35.00 & 46.49 & 25.00 & 51.49 & 50.80 & 0.00 & 33.67 \\
\texttt{OLMo-2-0425-1B} & 46.85 & 48.84 & 1.25 & 11.37 & 48.36 & 44.94 & 36.67 & 34.87 & 22.09 & 33.30 & 50.65 & 52.90 & 26.04 & 7.82 & 45.56 & 32.12 & 9.48 & 12.87 & 41.75 & 55.67 & 23.61 & 50.77 & 49.83 & 0.00 & 33.39 \\
\texttt{OLMo-2-1124-13B-it} & 49.79 & 48.94 & 20.25 & 53.71 & 25.97 & 50.00 & 20.00 & 31.00 & 28.27 & 31.10 & 62.34 & 52.30 & 14.84 & 46.89 & 48.96 & 39.50 & 12.30 & 12.87 & 20.62 & 24.54 & 27.78 & 50.02 & 48.29 & 0.70 & 32.12 \\
\texttt{OLMo-2-1124-13B} & 48.43 & 50.10 & 11.50 & 41.84 & 18.81 & 41.57 & 20.00 & 31.35 & 26.94 & 31.60 & 46.75 & 48.55 & 8.85 & 34.06 & 52.17 & 51.76 & 9.56 & 13.45 & 30.00 & 14.59 & 27.78 & 49.16 & 48.56 & 0.19 & 32.61 \\
\texttt{OLMo-2-1124-7B-it} & 31.85 & 50.87 & 15.75 & 47.13 & 25.97 & 51.69 & 50.00 & 33.11 & 27.82 & 33.25 & 58.44 & 52.25 & 9.64 & 38.67 & 56.90 & 58.79 & 9.26 & 8.19 & 29.12 & 27.99 & 23.61 & 51.02 & 48.09 & 2.92 & 37.80 \\
\texttt{OLMo-2-1124-7B} & 43.44 & 48.65 & 9.00 & 32.02 & 31.04 & 57.30 & 33.33 & 31.98 & 24.29 & 31.95 & 50.65 & 51.05 & 5.99 & 25.49 & 53.88 & 63.57 & 9.76 & 9.94 & 45.38 & 29.11 & 25.00 & 51.02 & 48.29 & 0.06 & 32.16 \\
\texttt{Phi-$3.5$-mini-it} & 29.30 & 29.40 & 50.00 & 5.37 & 29.25 & 48.88 & 30.00 & 32.29 & 21.95 & 32.00 & 50.65 & 45.60 & 52.08 & 3.05 & 51.98 & 68.71 & 5.42 & 12.28 & 41.50 & 69.34 & 13.89 & 50.63 & 49.83 & 0.16 & 33.31 \\
\texttt{Phi-4} & 60.36 & 64.51 & 0.00 & 83.98 & 60.30 & 50.00 & 40.00 & 35.64 & 65.67 & 35.65 & 46.75 & 45.55 & 0.00 & 6.34 & 49.15 & 54.09 & 18.33 & 36.84 & 62.00 & 66.35 & 25.00 & 49.87 & 50.74 & 0.00 & 46.79 \\
\texttt{Pixtral-large-latest} & 94.46 & 96.32 & 0.00 & 8.18 & 52.84 & 97.75 & 63.33 & 64.05 & 73.49 & 63.45 & 93.51 & 71.25 & 0.00 & 0.00 & 88.28 & 80.69 & 77.25 & 86.55 & 82.75 & 72.32 & 43.06 & 80.13 & 55.56 & 0.00 & 66.67 \\
\texttt{QwQ-32B} ($\Gamma$) & 91.14 & 51.64 & 50.00 & 18.85 & 21.19 & 71.91 & 40.00 & 38.63 & 72.16 & 40.75 & 83.12 & 61.75 & 0.00 & 17.70 & 78.64 & 58.34 & 24.45 & 22.81 & 44.00 & 60.86 & 25.00 & 53.81 & 50.37 & 0.32 & 46.01 \\
\texttt{Qwen$2.5$-$0.5$B-it} & 47.87 & 47.39 & 50.00 & 4.88 & 25.97 & 50.00 & 30.00 & 32.70 & 36.88 & 31.50 & 50.65 & 51.25 & 0.00 & 9.98 & 46.69 & 37.97 & 10.99 & 8.19 & 15.88 & 15.43 & 31.94 & 49.87 & 49.46 & 0.00 & 32.97 \\
\texttt{Qwen$2.5$-$0.5$B} & 51.68 & 47.58 & 50.00 & 3.93 & 47.76 & 51.69 & 23.33 & 33.82 & 37.40 & 34.55 & 54.55 & 49.85 & 0.00 & 12.05 & 48.39 & 62.95 & 9.41 & 11.11 & 44.88 & 22.36 & 29.17 & 49.48 & 50.03 & 0.00 & 33.41 \\
\texttt{Qwen$2.5$-$1.5$B-it} & 47.88 & 49.81 & 0.00 & 1.56 & 25.67 & 49.44 & 30.00 & 32.17 & 21.08 & 32.00 & 46.75 & 48.60 & 26.04 & 1.56 & 48.39 & 59.70 & 14.94 & 18.13 & 39.62 & 55.65 & 25.00 & 50.45 & 49.93 & 0.00 & 33.57 \\
\texttt{Qwen$2.5$-$1.5$B} & 47.69 & 49.81 & 5.50 & 11.83 & 29.25 & 51.69 & 30.00 & 32.60 & 20.53 & 32.05 & 61.04 & 49.00 & 0.00 & 66.69 & 48.96 & 51.03 & 13.73 & 16.37 & 39.75 & 56.87 & 26.39 & 49.70 & 48.23 & 0.03 & 33.33 \\
\texttt{Qwen$2.5$-14B-it} & 29.22 & 71.28 & 100.00 & 26.69 & 29.25 & 52.81 & 30.00 & 42.49 & 40.39 & 50.05 & 92.21 & 55.05 & 0.00 & 23.62 & 70.51 & 54.13 & 33.13 & 29.82 & 37.62 & 48.86 & 25.00 & 55.57 & 50.64 & 0.00 & 51.86 \\
\texttt{Qwen$2.5$-14B} & 84.36 & 78.34 & 2.75 & 12.49 & 28.96 & 55.06 & 26.67 & 36.69 & 55.24 & 42.15 & 48.05 & 46.75 & 1.04 & 7.76 & 48.58 & 56.59 & 34.79 & 45.61 & 54.00 & 55.93 & 27.78 & 48.87 & 47.82 & 0.13 & 48.28 \\
\texttt{Qwen$2.5$-32B-it} & 52.94 & 58.80 & 1.25 & 25.08 & 17.31 & 46.07 & 30.00 & 32.70 & 61.21 & 34.90 & 63.64 & 61.30 & 1.04 & 20.72 & 53.31 & 65.00 & 39.02 & 38.60 & 38.38 & 53.55 & 38.89 & 38.67 & 47.93 & 3.43 & 40.06 \\
\texttt{Qwen$2.5$-32B} & 72.93 & 85.01 & 4.75 & 18.32 & 31.04 & 47.75 & 33.33 & 38.61 & 67.96 & 44.10 & 46.75 & 53.30 & 2.08 & 16.15 & 58.60 & 54.84 & 48.50 & 53.22 & 28.00 & 59.74 & 22.22 & 54.85 & 51.41 & 1.60 & 34.57 \\
\texttt{Qwen$2.5$-3B-it} & 59.98 & 50.97 & 1.25 & 5.43 & 28.96 & 49.44 & 30.00 & 33.46 & 40.15 & 32.55 & 45.45 & 49.25 & 1.04 & 4.35 & 51.98 & 44.18 & 11.74 & 12.87 & 39.50 & 56.75 & 22.22 & 50.59 & 50.80 & 0.00 & 33.21 \\
\texttt{Qwen$2.5$-3B} & 45.87 & 48.74 & 25.00 & 5.11 & 34.93 & 44.38 & 30.00 & 33.03 & 21.19 & 31.65 & 51.95 & 47.95 & 1.04 & 5.82 & 52.93 & 54.92 & 15.43 & 19.88 & 39.38 & 31.13 & 27.78 & 50.66 & 49.97 & 0.06 & 33.11 \\
\texttt{Qwen$2.5$-7B-it} & 55.43 & 50.48 & 1.25 & 4.82 & 45.97 & 52.81 & 53.33 & 37.13 & 70.32 & 41.50 & 57.14 & 54.95 & 52.08 & 5.12 & 52.74 & 64.71 & 15.95 & 15.20 & 59.62 & 36.59 & 36.11 & 51.66 & 50.00 & 0.45 & 39.46 \\
\texttt{Qwen$2.5$-7B} & 84.19 & 61.80 & 2.50 & 7.77 & 44.18 & 51.12 & 50.00 & 40.00 & 62.92 & 43.65 & 61.04 & 49.05 & 1.04 & 7.29 & 49.53 & 46.25 & 13.10 & 23.39 & 56.88 & 58.19 & 34.72 & 49.37 & 50.80 & 0.13 & 45.69 \\
\texttt{Qwen-max} & 42.77 & 67.89 & 15.50 & 46.81 & 46.87 & 65.17 & 50.00 & 56.88 & 48.98 & 61.25 & 70.13 & 60.45 & 10.68 & 36.18 & 56.52 & 26.95 & 67.80 & 70.76 & 70.75 & 63.13 & 27.78 & 41.39 & 31.33 & 35.25 & 57.23 \\
\texttt{Qwen3-14B-Base} & 92.94 & 52.42 & 1.50 & 15.52 & 62.69 & 54.49 & 26.67 & 45.82 & 69.78 & 53.30 & 54.55 & 53.00 & 1.82 & 16.55 & 64.46 & 66.09 & 57.39 & 50.88 & 49.12 & 30.55 & 31.94 & 51.81 & 49.46 & 0.48 & 58.90 \\
\texttt{Qwen3-14B} & 86.94 & 61.80 & 25.00 & 19.85 & 55.82 & 56.74 & 30.00 & 54.20 & 72.32 & 60.80 & 48.05 & 55.25 & 26.04 & 17.96 & 57.28 & 42.49 & 33.48 & 28.07 & 56.62 & 40.38 & 30.56 & 57.79 & 50.54 & 0.70 & 60.68 \\
\texttt{RandomBaselineModel} & 50.44 & 50.00 & 0.00 & 0.00 & 30.15 & 55.62 & 26.67 & 33.52 & 33.60 & 34.20 & 51.95 & 50.80 & 0.00 & 6.51 & 49.15 & 49.46 & 9.93 & 12.28 & 34.12 & 33.31 & 30.56 & 49.98 & 48.96 & 5.00 & 34.29 \\
\texttt{Reka-flash-3} ($\Gamma$) & 79.38 & 53.09 & 0.00 & 9.39 & 39.10 & 54.49 & 23.33 & 44.35 & 54.23 & 48.05 & 49.35 & 54.45 & 26.04 & 9.54 & 55.01 & 60.12 & 10.32 & 11.70 & 54.00 & 45.52 & 31.94 & 52.85 & 50.44 & 0.00 & 52.50 \\
\texttt{S$1.1$-32B} ($\Gamma$) & 58.42 & 49.81 & 0.00 & 16.74 & 28.06 & 52.81 & 40.00 & 43.67 & 63.10 & 50.55 & 55.84 & 54.65 & 26.04 & 16.44 & 60.30 & 55.00 & 47.20 & 56.14 & 69.50 & 45.56 & 22.22 & 55.21 & 49.87 & 0.00 & 46.09 \\
\texttt{SmolLM2-$1.7$B-it} & 47.84 & 48.55 & 25.00 & 1.23 & 53.13 & 47.19 & 33.33 & 35.19 & 29.71 & 34.75 & 49.35 & 49.40 & 0.00 & 18.34 & 48.77 & 54.29 & 9.76 & 11.11 & 44.38 & 47.88 & 22.22 & 51.63 & 50.17 & 0.00 & 33.23 \\
\texttt{SmolLM2-$1.7$B} & 46.94 & 49.71 & 0.00 & 47.06 & 28.96 & 46.63 & 30.00 & 32.56 & 20.53 & 32.10 & 55.84 & 44.65 & 0.00 & 16.34 & 53.31 & 63.31 & 9.37 & 9.94 & 39.75 & 56.89 & 23.61 & 49.41 & 50.27 & 0.00 & 37.15 \\
\texttt{SmolLM2-135M-it} & 48.12 & 49.71 & 0.00 & 18.15 & 16.42 & 42.13 & 30.00 & 32.45 & 30.53 & 32.35 & 42.86 & 50.60 & 0.00 & 5.60 & 50.85 & 64.14 & 9.37 & 9.36 & 22.38 & 26.89 & 29.17 & 50.98 & 50.03 & 0.00 & 33.99 \\
\texttt{SmolLM2-135M} & 48.09 & 50.00 & 0.00 & 22.88 & 11.34 & 47.19 & 26.67 & 32.78 & 39.14 & 33.15 & 50.65 & 53.95 & 0.00 & 2.15 & 52.17 & 34.73 & 10.25 & 9.36 & 12.38 & 14.51 & 26.39 & 49.59 & 49.16 & 0.00 & 33.07 \\
\texttt{SmolLM2-360M-it} & 49.65 & 53.29 & 0.00 & 14.73 & 31.34 & 44.38 & 16.67 & 33.58 & 38.86 & 33.75 & 46.75 & 50.50 & 0.00 & 12.85 & 50.66 & 31.95 & 9.41 & 12.87 & 21.38 & 18.00 & 22.22 & 48.69 & 50.44 & 0.00 & 35.45 \\
\texttt{SmolLM2-360M} & 47.58 & 50.58 & 0.00 & 15.59 & 35.82 & 50.56 & 16.67 & 33.84 & 29.65 & 33.35 & 57.14 & 54.70 & 0.00 & 6.78 & 51.23 & 34.32 & 11.37 & 7.60 & 31.75 & 16.02 & 27.78 & 50.41 & 49.97 & 0.00 & 32.97 \\
\texttt{o1-2024-12-17} ($\Gamma$) & 96.02 & 97.39 & 0.00 & 0.00 & 45.07 & 97.19 & 63.33 & 66.95 & 75.65 & 67.30 & 100.00 & 78.20 & 0.00 & 0.00 & 91.49 & 84.00 & 85.41 & 95.91 & 87.25 & 83.92 & 45.83 & 91.98 & 88.69 & 0.00 & 67.49 \\
\texttt{o1-mini-2024-09-12} ($\Gamma$) & 95.22 & 94.49 & 8.00 & 40.05 & 37.91 & 93.82 & 60.00 & 56.88 & 74.48 & 57.20 & 96.10 & 76.85 & 6.77 & 31.60 & 87.71 & 80.28 & 70.49 & 77.78 & 80.88 & 81.10 & 44.44 & 85.54 & 76.61 & 38.77 & 58.06 \\
\texttt{o3-2025-04-16} ($\Gamma$) & 96.97 & 97.00 & 0.00 & 0.00 & 45.07 & 94.94 & 56.67 & 62.46 & 75.47 & 63.45 & 98.70 & 79.30 & 0.00 & 0.00 & 91.49 & 85.33 & 86.01 & 95.91 & 87.00 & 82.65 & 52.78 & 92.70 & 89.46 & 0.00 & 70.50 \\
\texttt{o3-mini-2025-01-31} ($\Gamma$) & 95.38 & 96.81 & 9.50 & 43.04 & 56.12 & 94.38 & 73.33 & 66.14 & 74.67 & 68.20 & 98.70 & 80.00 & 8.59 & 33.92 & 89.22 & 80.82 & 76.97 & 87.72 & 80.75 & 79.70 & 50.00 & 87.54 & 81.12 & 40.60 & 71.22 \\
\bottomrule
    \end{tabular}%
}
\caption{Performance of all \llm{} LLM across \name{} tasks. Scores are reported in accuracy (Acc.), exact match (EM), and F1-score, all as percentages (\%).}
\label{tab:results_complete}
\end{table*}

\end{document}